\documentclass{article} 
\usepackage{iclr2026_conference,times}
\iclrfinalcopy

\usepackage{xcolor} 

\usepackage{amsmath,amsfonts,bm}

\def\eqref#1{equation~\ref{#1}}

\def\1{\bm{1}}

\DeclareMathAlphabet{\mathsfit}{\encodingdefault}{\sfdefault}{m}{sl}
\SetMathAlphabet{\mathsfit}{bold}{\encodingdefault}{\sfdefault}{bx}{n}

\usepackage{algorithm}
\usepackage{tabularx}

\usepackage{algpseudocode}   
\usepackage{hyperref}
\usepackage{url}
\usepackage[table]{xcolor}
\usepackage{subcaption}   
\usepackage{ragged2e}     
\usepackage{tcolorbox}    
\usepackage{tabularx,booktabs,array,ragged2e}

\newcommand{\methodcard}[3]{%
  \textbf{#1} \, (\textit{tokens:} #2)\par
  \medskip
  \fbox{\parbox{\linewidth}{\small #3}}%
}

\newcommand{\casehdr}[2]{%
  \small\textbf{#1}\par
  \textit{#2}\par\medskip
}

\newcolumntype{Y}{>{\RaggedRight\arraybackslash}X}

\usepackage{xcolor} 
\usepackage{bigstrut}
\usepackage{multirow}
\usepackage{graphicx}
\usepackage{amsmath}
\usepackage{enumitem}

\newcommand{\tool}{\textsc{FR-Ponder }}

\title{Learning to Ponder: Adaptive Reasoning in Latent Space}

\author{
Yixin He$^*$ \\
University of Southern California \\
\texttt{yixinhe@usc.edu}
\And
Lumingyuan Tang$^*$ \\
Independent Researcher \\
\texttt{tanglujay@gmail.com}
\thanks{These authors contributed equally}
}

\begin{document}

\maketitle

\begin{abstract}

Test-time compute has emerged as a key paradigm for enhancing LLM reasoning, yet prevailing approaches like Best-of-$N$ and majority voting apply uniform depth across inputs, wasting computation on simple queries while potentially under-thinking complex ones.  We present FR‑Ponder, a single‑graph, backbone‑training‑free framework that allocates instance‑adaptive reasoning compute via latent steering. A  less than 1M‑param controller observes hidden states and decides to halt or apply a small ponder step by adding a pre‑computed steering vector to frozen representations. Our method extracts the latent steering vector associated with deeper reasoning outputs and direct IO from LLM and re-applies it through a tunable scaling factor, allowing the model to adapt its reasoning depth to the complexity of each input. To balance performance and computational cost, we employ Group Relative Policy Optimization (GRPO) as a reward signal to adaptively regulate reasoning depth, achieving task accuracy while mitigating overreasoning. Through curriculum learning and careful reward engineering, FR-Ponder learns calibrated compute allocation correlated with problem difficulty. On GSM8K and MATH500, FR‑Ponder improves the compute–accuracy frontier, delivering lower FLOPs with better matched accuracy and comparing favorably to early‑exit baselines, without modifying backbone weights. Analyses visualize interpretable steering directions and show learned compute allocation correlates with problem difficulty.

\end{abstract}

\section{Introduction}

Large language models (LLMs) have achieved remarkable success across diverse reasoning tasks, yet they exhibit a fundamental inefficiency: \emph{fixed computational allocation}. Whether processing a simple factual query or solving a complex mathematical problem, current LLMs expend identical compute per token~\citep{kaplan2020scaling,hoffmann2022training}. This rigid approach leads to systematic over-computation on easy instances and under-allocation on challenging ones, creating a compute-accuracy mismatch that becomes increasingly problematic as models scale to hundreds of billions of parameters~\citep{touvron2023llama2,achiam2023gpt4}.

Recent efforts to address this inefficiency fall into three categories, each with significant limitations. \emph{Multi-pass methods} like chain-of-thought prompting~\citep{wei2022chain} and self-consistency~\citep{wang2022selfconsistency} achieve adaptive reasoning by sampling multiple trajectories, but multiply inference costs by the number of passes. \emph{Architectural modifications} including early-exit mechanisms~\citep{schuster2022confident,zhou2020bert} and layer skipping~\citep{elhoushi2024layerskip} require model retraining, limiting deployment flexibility and often degrading base model capabilities. \emph{Speculative decoding} approaches~\citep{leviathan2023fast,chen2023accelerating} accelerate inference through draft-verify paradigms but require maintaining multiple models and provide only coarse-grained adaptation.

The recent \emph{Fractional Reasoning} framework~\citep{liu2025fractional} introduced a promising direction: extracting ``reasoning vectors'' from contrastive prompts and applying them with tunable intensity to control reasoning depth. However, this approach requires manual tuning of the scaling factor $\alpha$ for each problem type, lacks dynamic adaptation within a single inference, and provides no principled method for learning optimal compute allocation.

To address this, we introduce \textbf{\tool} (\textbf{F}ractional \textbf{R}eason Ponder Framework), a framework that transforms inference depth into a \emph{learnable decision process}. As shown in Figure~\ref{fig:overview}, we provide an overview of our method \tool.
  Our key insight is decomposing adaptive computation into two orthogonal problems: (1) \emph{what to think about} via steering vectors encoding reasoning directions, and (2) \emph{how long to think} via a lightweight pondering controller. At each decoding step, \tool observes the current hidden state and either halts to emit a token or applies an additive ``thought step'' along learned steering vectors:

\begin{equation}
z_{k+1} = z_k + \phi(z_k) \cdot \Delta z(h_{\text{steer}}), \quad \text{halt if } \phi(z_k) \leq \tau
\end{equation}
where $\phi(\cdot)$ is a learned pondering probability, $h_{\text{steer}}$ represents steering vectors extracted via contrastive activation~\citep{zou2023representation,rimsky2024steering}, and $\tau$ is a halting threshold.

This design enables several critical advantages over prior work:

\textbf{One-pass, zero-backbone-finetune operation.} Unlike methods requiring multiple forward passes or model retraining, \tool operates in a single inference pass with the base LLM completely frozen. Only a small controller network ($\leq 1$M parameters) is trained, preserving the model's original capabilities while adding less than 0.01\% parameter overhead.

\textbf{Fine-grained, instance-adaptive depth.} Rather than applying uniform depth across all tokens or problems, \tool makes per-token pondering decisions based on the evolving hidden state. This creates a continuous spectrum of reasoning intensity that automatically adapts to local complexity—spending more compute on challenging reasoning steps while quickly resolving simple continuations.

\textbf{Multi-objective reward.} We formulate adaptive computation as a reinforcement learning problem with carefully designed rewards that balance multiple objectives:
\begin{equation}
R = w_{\text{acc}} \cdot \text{Accuracy} - w_{\text{flops}} \cdot \text{FLOPs} + w_{\text{comp}} \cdot \text{Completeness} + w_{\text{qual}} \cdot \text{Quality} - w_{\text{rep}} \cdot \text{Repetition}
\end{equation}
This rich reward signal addresses critical challenges in adaptive reasoning, including partial credit for mathematical solutions, anti-repetition mechanisms to prevent pondering collapse, and completeness validation to ensure full reasoning traces.

\textbf{Variance-reduced policy optimization.} We train the pondering controller using Group Relative Policy Optimization (GRPO)~\citep{shao2024deepseekmath}, a value-free policy gradient method that achieves variance reduction through in-group baselines. By sampling multiple trajectories per input and using group-average rewards as baselines, GRPO provides stable learning without requiring a separate value network—crucial for our lightweight controller design.

Our contributions are:
\begin{itemize}
\item \textbf{Conceptual:} We formulate adaptive inference as a meta-cognitive decision process, where the model learns to allocate computation based on evolving internal states rather than external heuristics.
\item \textbf{Technical:} We develop a complete framework combining steering vector extraction, lightweight pondering control, multi-component reward engineering, and curriculum-based training that achieves stable learning of adaptive policies.
\item \textbf{Empirical:} \tool achieves 30--50\% token reduction on GSM8K, MATH500 and GPQA while maintaining or improving accuracy, with analysis revealing calibrated halting patterns and interpretable steering directions.
\end{itemize}


\section{Related Work}

\subsection{Chain-of-Thought Reasoning}
Chain-of-Thought (CoT) reasoning enhances large language models (LLMs) by generating structured intermediate steps before final outputs. By decomposing complex tasks into sequential reasoning chains, CoT can be elicited via prompting \cite{wei2022chain, wang2022towards, qin2023crosslingual}, supervised fine-tuning \cite{kojima2022large, yu2025longshort, byun2024ares}, or reinforcement learning \cite{lightman2023let, shen2025satori, xie2025interleaved}. Theoretical analyses show that feeding intermediate outputs back as inputs effectively deepens transformers, increasing expressivity and enabling more sophisticated inference \cite{feng2023towards, li2024empowers, merrill2025logdepth}. Structuring tokens as symbols, patterns, and text further produces concise and efficient reasoning chains \cite{zhang2023language, pan2023logiclm, doll2025emergent}.

Beyond standard CoT, techniques to improve robustness and fidelity include reward-based frameworks for selective retention or reranking \cite{jiang2025eorm, rankcot2025, scan2024rrm}, self-consistency methods for evaluating agreement across sampled chains \cite{wang2022selfconsistency, yang2023selfagreement, surveyprompt2025}, and iterative refinement through self-correction or reflection-based prompting \cite{madaan2023selfrefine, xue2023rcot, iort2025}. However, the autoregressive nature of CoT limits its ability to emulate human-like planning in complex tasks \cite{wu2025depth, hao2024coconut, treeofthoughts2024}. Some approaches mitigate this by integrating explicit search procedures or training on search trajectories \cite{wu2025depth, hao2024coconut}, while recent work indicates that latent-space reasoning can spontaneously produce patterns resembling breadth-first search without supervision \cite{hao2024coconut, system15reasoning2025}.

\subsection{Latent Reasoning}

Latent reasoning reflects the internal computations of large language models (LLMs) within hidden representations, which may diverge from explicit Chain-of-Thought (CoT) outputs. Prior work shows that intermediate reasoning variables can often be recovered from hidden states \cite{bharadwaj2024understanding, stateshidden2024, reasoning_models_selfverification2025}, and targeted interventions on these representations can modulate model behavior \cite{wang2024sadi, zhufabi2025expertsteer, su2025activationsteering}. Multiple latent reasoning paths suggest that LLMs employ diverse internal strategies independent of token-level outputs, revealing inherent unfaithfulness between latent and explicit reasoning \cite{yee2024dissociation, li2024bridgingreasoners, bharadwaj2024understanding}.

Several strategies aim to enhance latent reasoning. Training with learnable or filler tokens improves performance on parallelizable tasks \cite{pfau2024thinkdot, coconut2024, latentreasoning2025}, while discrete planning tokens guide subsequent reasoning steps \cite{wang2023planningtokens, coconut2024, planningtokens2025}. Knowledge distillation and progressive curricula internalizing CoT enable complex reasoning within latent space \cite{system15reasoning2025, latentreasoning2025, coconut2024}, and architectures like looped transformers exploit iterative feedback of internal states \cite{yang2023loopedtransformers, transformerfam2024, system15reasoning2025}. Despite these advances, latent reasoning remains less interpretable than CoT, and generalization to complex tasks is still an open challenge \cite{system15reasoning2025, latentreasoning2025, coconut2024}.

\subsection{Generalized Reinforcement Learning for Reasoning}

Reinforcement learning (RL) has been applied to enhance reasoning in large language models (LLMs), primarily through policy optimization. Early work such as DeepSeek-R1 employs Group Relative Policy Optimization (GRPO) to encourage multi-step reasoning \cite{deepseek_r1_grpo, deepseek_r1_training}. While Proximal Policy Optimization (PPO) improves response length and task performance, it suffers from high sample complexity due to repeated rollouts \cite{ppo_sample_complexity}. 

GRPO and its extensions address these limitations by improving efficiency and convergence. DAPO (Decoupled Clip and Dynamic Sampling Policy Optimization) stabilizes training via reward shaping, dynamic sampling, and token-level gradients \cite{yu2025dapo}, while Dr. GRPO (GRPO Done Right) mitigates optimization biases by adjusting reward normalization and length bias \cite{liu2025drgrpo}. Collectively, these methods show that GRPO provides a principled and efficient framework for adaptive reasoning in LLMs, balancing stability, scalability, and performance.

\section{Methodology}



Traditional approaches to adaptive computation either require expensive architectural modifications \cite{graves2016adaptive} or joint training with the base model \cite{elbayad2020depth}. In contrast, \tool introduces a lightweight pondering controller that operates in the latent space of frozen pre-trained models, making dynamic halting decisions based on steered representations. This controller is trained via Group Relative Policy Optimization with a carefully designed multi-component reward function that balances accuracy and computational efficiency.

\begin{figure}[t]
  \centering
  \includegraphics[width=\linewidth]{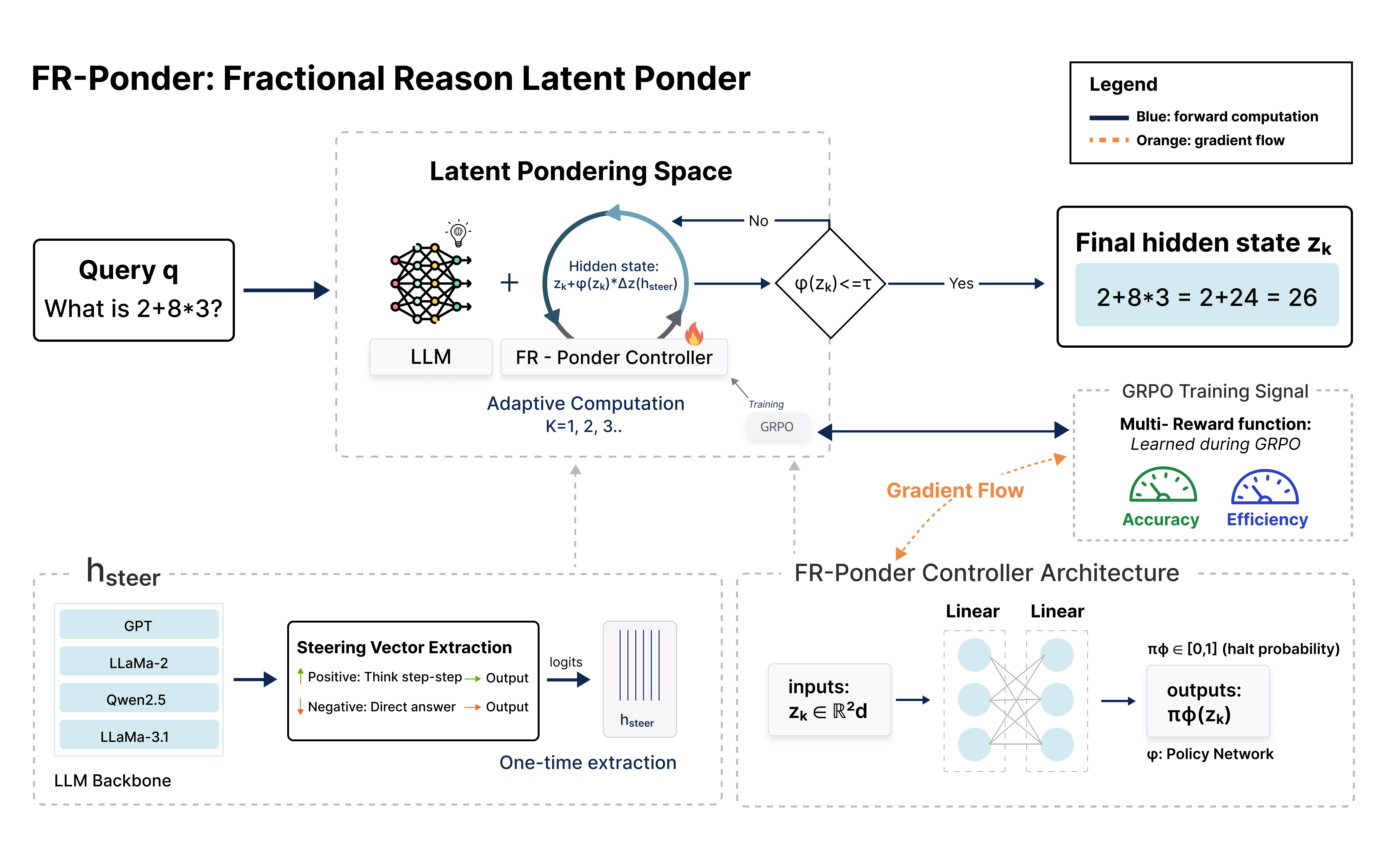}
    \caption{\textbf{Method overview of \tool.}
    Given a query $q$, a \textbf{frozen} LLM produces the initial hidden state $z_0$ at the final token position.
    Inside the \emph{Latent Pondering Space}, a lightweight controller $\varphi_\phi$ reads $z_k$ and decides whether to halt; if $\varphi_\phi(z_k) \le \tau$ the process stops, otherwise a pondering step is applied:
    $z_{k+1} = z_k + \alpha_k \, h_{\text{steer}}$ (optionally with a layer schedule), repeated for $k=1,\dots,K$.
    The final state $z_K$ is decoded to the output answer.
    The steering vector $h_{\text{steer}}$ is \emph{extracted once} via contrastive prompts (``think step-by-step'' vs. ``direct answer'') and kept fixed thereafter.
    During training, \emph{only} the controller is updated (orange dashed arrows) with GRPO using a multi-component reward that primarily balances accuracy and FLOPs; the backbone and $h_{\text{steer}}$ remain frozen.
    The controller is a $\leq 1$M-parameter MLP.
    Blue arrows denote the single-pass forward/inference path; orange arrows denote gradient flow. \label{fig:overview}}
\end{figure}

The overview of \tool is presented in Fig. \ref{fig:overview}. The fundamental insight driving our approach is that reasoning depth can be modulated through controlled perturbations in representation space, guided by steering vectors that encode the difference between deliberative and direct reasoning modes. This enables the model to adaptively "think longer" on complex problems while maintaining efficiency on simpler ones, all without modifying the original model parameters.

We present \tool, a framework for adaptive inference depth in large language models that achieves single-pass, backbone-training-free deployment while maintaining superior compute-accuracy trade-offs. The core innovation lies in treating adaptive computation as a meta-cognitive process where the model learns when to allocate additional computational resources during inference.

\subsection{Problem Formulation}

The central challenge in adaptive inference is determining the optimal amount of computation to allocate for each input while maintaining both accuracy and efficiency. We formulate this as a sequential decision-making problem where, at each reasoning step, an agent must choose between continuing computation (potentially improving accuracy) or halting to produce an answer (saving computational resources).

This naturally leads to a Markov Decision Process (MDP) \cite{puterman1990markov} formulation, which provides a principled framework for modeling sequential decision-making under uncertainty. The MDP framework is particularly well-suited for our setting because: (1) the decision at each step depends only on the current representation state (Markov property), (2) the agent receives rewards that balance accuracy and efficiency, and (3) the finite-horizon nature ensures bounded computation.

Let $\mathcal{M}_\theta$ denote a frozen pre-trained language model with parameters $\theta$, and let $x \in \mathcal{X}$ be an input sequence. The model processes the input and produces an initial hidden state $\mathbf{z}_0 \in \mathbb{R}^d$ at the final token position. This state $\mathbf{z}_0$ serves as the starting point for our adaptive pondering process, containing the model's initial understanding of the problem.

We define the MDP tuple $(\mathcal{S}, \mathcal{A}, \mathcal{T}, \mathcal{R}, \gamma)$ where:

\begin{itemize}
\item $\mathcal{S} = \mathbb{R}^d$ represents the state space of hidden representations. Each state $\mathbf{z}_k \in \mathcal{S}$ encodes the model's current understanding after $k$ pondering steps. The choice of the full hidden representation space allows for rich state representations that can capture subtle differences in reasoning progress.

\item $\mathcal{A} = \{0, 1\}$ denotes the binary action space where $a=0$ means "halt and produce answer" and $a=1$ means "continue pondering." This binary formulation simplifies the decision space while capturing the essential trade-off between accuracy and efficiency.

\item $\mathcal{T}: \mathcal{S} \times \mathcal{A} \rightarrow \mathcal{S}$ defines state transitions via steering vector application. When $a=1$ (continue), the transition applies steering: $\mathbf{z}_{k+1} = \mathcal{T}(\mathbf{z}_k, 1) = \mathbf{z}_k + \alpha_k \mathbf{h}_{\text{steer}}$. When $a=0$ (halt), the state remains unchanged. This deterministic transition function ensures reproducible behavior while allowing controlled exploration of the representation space.

\item $\mathcal{R}: \mathcal{S} \times \mathcal{A} \rightarrow \mathbb{R}$ specifies the multi-component reward function that balances multiple objectives including accuracy, computational efficiency, reasoning completeness, output quality, and anti-repetition measures. This comprehensive reward design prevents the agent from optimizing for a single metric at the expense of others.

\item $\gamma = 1$ indicates an undiscounted, finite-horizon problem. The undiscounted formulation is appropriate because we care equally about all steps in the reasoning process, and the finite horizon (maximum $K$ steps) ensures bounded computation.
\end{itemize}

Our objective is to learn a policy $\pi_\phi: \mathcal{S} \rightarrow [0,1]$ parameterized by $\phi$ that outputs the probability of continuing computation. The policy must balance two competing objectives: maximizing task performance while minimizing computational cost. This leads to the following optimization problem:

\begin{equation}
\max_\phi \mathbb{E}_{\tau \sim \pi_\phi} \left[ R(\tau) - \lambda \cdot \text{FLOPs}(\tau) \right]
\label{eq:objective}
\end{equation}

This objective function \eqref{eq:objective} encodes the fundamental trade-off in adaptive inference. The first term $R(\tau)$ captures the quality of the reasoning process and final answer, incorporating multiple dimensions of performance. The second term $\lambda \cdot \text{FLOPs}(\tau)$ penalizes computational overhead, where $\text{FLOPs}(\tau) = \sum_{k=0}^T c_k$ represents the cumulative floating-point operations across the trajectory $\tau = \{(\mathbf{z}_k, a_k)\}_{k=0}^T$. The hyperparameter $\lambda$ controls the strength of the efficiency constraint—larger values of $\lambda$ encourage shorter reasoning sequences, while smaller values allow more extensive deliberation.

The expectation is taken over trajectories $\tau$ generated by policy $\pi_\phi$, which induces a distribution over reasoning lengths and paths through the representation space. This formulation naturally handles the stochastic nature of the pondering process while providing clear gradients for policy optimization.

\subsection{Adaptive Pondering Mechanism}

The adaptive pondering mechanism is the core component that enables \tool to dynamically allocate computation based on problem complexity. Unlike fixed-depth approaches, our mechanism allows the model to continue refining its internal representations until it reaches sufficient confidence or exhausts the computational budget. This section describes how representations evolve during pondering and how the controller makes halting decisions.

\paragraph{State Evolution Dynamics}

The evolution of hidden states during pondering is governed by controlled applications of steering vectors. At each pondering step $k$, we apply a carefully calibrated perturbation to guide the representation toward more deliberative reasoning states. This process must be stable (preventing divergence) while still allowing meaningful exploration of the representation space.

At each pondering step $k$, the system evolves the latent state through additive steering:

\begin{equation}
\mathbf{z}_{k+1} = \mathbf{z}_k + \alpha(k) \cdot \mathbf{h}_{\text{steer}}^{(\ell_k)}
\label{eq:evolution}
\end{equation}

This equation \eqref{eq:evolution} defines the core dynamics of our pondering process. The additive update $\mathbf{z}_k + \alpha(k) \cdot \mathbf{h}_{\text{steer}}^{(\ell_k)}$ progressively shifts the hidden representation in the direction of deliberative reasoning. The scaling factor $\alpha(k)$ controls the magnitude of each update, while the steering vector $\mathbf{h}_{\text{steer}}^{(\ell_k)}$ determines the direction.

The time-dependent scaling factor $\alpha(k) = \alpha_0 \cdot e^{-\beta k}$ implements exponential decay to prevent unbounded growth and ensure convergence. This decay serves several important purposes: (1) it provides larger updates early in the pondering process when the representation may be far from optimal, (2) it ensures smaller, more refined updates as pondering progresses, and (3) it guarantees bounded total displacement from the original representation.

The layer index $\ell_k$ can vary across steps for multi-layer steering, allowing the pondering process to engage different levels of representation. Early steps might apply steering at higher layers to modify high-level reasoning patterns, while later steps might focus on lower layers to refine specific details.

The exponential decay ensures mathematical stability:

\begin{equation}
\label{eq:bounded_evolution_discrete}
\bigl\|\mathbf{z}_T-\mathbf{z}_0\bigr\|_2
= \Bigl\| \sum_{k=0}^{T-1} \alpha_0 e^{-\beta k}\,\mathbf{h}_{\text{steer}} \Bigr\|_2
\le \alpha_0\,\frac{1-e^{-\beta T}}{1-e^{-\beta}} \,\bigl\|\mathbf{h}_{\text{steer}}\bigr\|_2 .
\end{equation}

This bound \eqref{eq:bounded_evolution_discrete} is derived by summing the geometric series of decay factors. It guarantees that the total deviation $\|\mathbf{z}_T - \mathbf{z}_0\|_2$ from the original representation is bounded by $\frac{\alpha_0}{\beta} \cdot \|\mathbf{h}_{\text{steer}}\|_2$, regardless of the number of pondering steps $T$. This is crucial for maintaining the model's general capabilities while allowing controlled exploration of the reasoning space. The bound becomes tighter as $\beta$ increases, providing a tunable parameter for controlling the exploration extent.

\paragraph{Pondering Controller Architecture}

The pondering controller is the decision-making component that determines whether to continue pondering or halt at each step. The controller must be lightweight enough to add minimal computational overhead while being expressive enough to capture complex patterns in the representation space that indicate optimal stopping points.

The controller $f_\phi: \mathbb{R}^d \rightarrow [0,1]$ is implemented as a shallow neural network designed for computational efficiency and stable training. The architecture is carefully designed to balance expressiveness with efficiency, using only a few layers to minimize computational overhead while maintaining sufficient capacity to learn complex stopping policies.

\begin{align}
\mathbf{g}^{(0)} &= \text{LayerNorm}(\mathbf{z}_k) \label{eq:controller_norm}\\
\mathbf{g}^{(i)} &= \sigma_i(\mathbf{W}^{(i)}\mathbf{g}^{(i-1)} + \mathbf{b}^{(i)}), \quad i \in \{1, 2\} \label{eq:controller_hidden}\\
\pi_\phi(\mathbf{z}_k) &= \text{sigmoid}\left(\frac{\mathbf{w}^T\mathbf{g}^{(2)} + b}{\tau_{\text{temp}}}\right) \label{eq:controller_output}
\end{align}

The LayerNorm operation in equation \eqref{eq:controller_norm} serves as input preprocessing, normalizing the hidden state $\mathbf{z}_k$ to have zero mean and unit variance across the hidden dimension. This normalization is crucial for several reasons: (1) it stabilizes training by preventing gradient explosion/vanishing, (2) it makes the controller robust to the absolute scale of hidden representations, which can vary across different models and layers, and (3) it ensures that the controller focuses on the relative patterns in the representation rather than absolute magnitudes.

The hidden layers in equation \eqref{eq:controller_hidden} use standard fully connected transformations with nonlinear activations $\sigma_i \in \{\text{GELU}, \text{ReLU}\}$. We use only two hidden layers to maintain computational efficiency while providing sufficient capacity for learning complex decision boundaries. The weight matrices $\mathbf{W}^{(i)} \in \mathbb{R}^{h_i \times h_{i-1}}$ and bias vectors $\mathbf{b}^{(i)} \in \mathbb{R}^{h_i}$ are learned parameters, where $h_0 = d$ (input dimension), $h_1 = h_2 = 512$ (hidden dimensions), and the final layer maps to a scalar.

The output layer in equation \eqref{eq:controller_output} produces the continuation probability $\pi_\phi(\mathbf{z}_k) \in [0,1]$. The sigmoid function ensures the output is a valid probability, while the temperature parameter $\tau_{\text{temp}}$ controls the sharpness of the decision boundary. A lower temperature (e.g., $\tau_{\text{temp}} = 0.1$) produces more decisive, near-binary decisions, while a higher temperature (e.g., $\tau_{\text{temp}} = 1.0$) allows for more nuanced probability distributions. This temperature scaling is particularly important during curriculum learning, where we gradually transition from soft to hard decision boundaries.

The controller contains $|\phi| \leq 10^6$ parameters, representing less than $0.01\%$ overhead for billion-parameter models. This minimal parameter count is achieved through the shallow architecture and moderate hidden dimensions (512 units per layer). The low overhead ensures that \tool can be applied to large models without significant computational cost increases.

\textbf{Theorem 2} (Universal Approximation). For any continuous continuation value function $V^*$ on a compact subset $K \subset \mathbb{R}^d$, there exists a controller network $\phi_\theta$ with $O(\epsilon^{-d/s})$ parameters that satisfies $\sup_{\mathbf{z} \in K} |\phi_\theta(\mathbf{z}) - V^*(\mathbf{z})| \leq \epsilon$, where $s > 0$ depends on the smoothness of $V^*$.

\paragraph{Controlled Diffusion Analysis}

The pondering evolution can be modeled as a controlled diffusion process:
\begin{equation}
d\mathbf{Z}_t = \alpha(t) \mathbf{h}_{\text{steer}} dt + \sigma dW_t
\end{equation}

where the discrete implementation uses bounded additive updates with noise:
\begin{equation}
\mathbf{z}_{k+1} = \mathbf{z}_k + \alpha_k \cdot \mathbf{h}_{\text{steer}} + \boldsymbol{\xi}_k
\end{equation}

\textbf{Lemma 2} (Convergence to Stationary Distribution). Under Lipschitz conditions on the steering function, the discrete process converges to a unique stationary distribution $\mu^*$ with convergence rate $O(k^{-1/2})$.


\textbf{Theorem 3} (Overhead Bound). For input length $n$, model dimension $d$, and maximum pondering steps $K$, \tool's overhead is:
\begin{equation}
O(K \cdot (d + |\phi|)) = O(K \cdot d)
\end{equation}
Since $K \ll n$ and $|\phi| \ll d$, the relative overhead is $O(K/n) \approx O(1/\sqrt{n})$ for typical sequences.

\section{Experiments}

\subsection{Evaluation Setup}
\paragraph{Dataset}
We evaluate our models on three widely used reasoning benchmarks. \textbf{Math500} covers middle- and high-school tasks in algebra, geometry, and symbolic manipulation \cite{openaimath2024}. \textbf{GPQA} provides graduate-level scientific questions demanding both domain knowledge and logical inference \cite{rein2023gpqa}. \textbf{GSM8K} consists of grade-school math word problems and serves as a standard benchmark for arithmetic reasoning \cite{cobbe2021gsm8k}. Collectively, these datasets span elementary to expert-level challenges, providing a comprehensive testbed for evaluating adaptive reasoning.

\paragraph{Backbone LLMs}
We evaluate our methods across five state-of-the-art large language models to ensure robustness and generality. The LLaMA-3 series includes \textbf{LLaMA-3-8B-Instruct} and \textbf{LLaMA-3-70B-Instruct}, representing mid- and large-scale transformer architectures optimized for instruction-following tasks \cite{meta2025llama3}. The Qwen-2.5 series comprises \textbf{Qwen-2.5-0.5B-Instruct}, \textbf{Qwen-2.5-3B-Instruct}, and \textbf{Qwen-2.5-7B-Instruct}, offering a spectrum of model capacities for evaluating performance scaling \cite{qwen2025qwen2.5}. This diverse selection allows us to assess adaptive reasoning across models of varying sizes and architectures, providing insights into both efficiency and effectiveness.

\paragraph{Comparison Baselines}

We evaluate our approach against four competitive baselines in mathematical reasoning tasks:

\begin{itemize}
    \item \textbf{Direct (Non-CoT).} The model generates outputs directly, without intermediate reasoning steps or structured prompting.

    \item \textbf{Chain-of-Thought (CoT)} CoT prompting elicits step-by-step reasoning. Multiple reasoning chains are sampled per input, with the final output selected as the model's answer .

    \item \textbf{Best-of-N (BoN).} This method samples N candidate outputs for each query and selects the highest-quality response using an external judge model, typically a pretrained LLM .

\end{itemize}

\paragraph{Evaluation Metrics}

We assess model performance using three complementary metrics that quantify both effectiveness and computational efficiency. \textbf{Accuracy (Exact Match)} denotes the proportion of model outputs that precisely correspond to the ground-truth solutions, providing a direct measure of problem-solving correctness, with higher values indicating superior performance \cite{}. \textbf{Average Tokens} represents the mean number of tokens generated per query, serving as a proxy for reasoning verbosity and computational overhead, where lower values signify more succinct and resource-efficient outputs \cite{}. \textbf{Average FLOPs} quantifies the mean floating-point operations expended per query, capturing inference-level computational cost, with lower values reflecting greater efficiency and reduced resource utilization \cite{}.

\paragraph{Implementation Details}

Without loss of generality, all baselines are evaluated under a unified experimental configuration. The maximum generation length is fixed at 500 newly generated tokens (excluding input text), with a batch size of 16. Decoding employs a temperature of 0.7. All experiments are conducted on NVIDIA RTX 6000 GPUs. For prompting, CoT adopts the instruction ``Please solve the math problem step by step,'' concatenated with the input question. All other methods, including \texttt{Best-of-N} and latent reasoning variants, use the simpler form ``Please solve the math problem,'' followed by the question.  Regarding method-specific settings, Best-of-N applies a majority-vote strategy, selecting the most frequently generated (i.e., most consistent) answer across samples. Following extensive empirical evaluation across candidate values, we adopt a termination threshold of 0.2 for \tool, which consistently yields strong performance by halting generation when the predicted next-token probability falls below this value.

\subsection{Main Results}
\begin{table}[htbp]
  \centering
  \caption{
Main evaluation results of different baseline methods and our \tool (FP) variants on three reasoning benchmarks: GSM8k, Math500, and GPQA. 
We report \textbf{ACC} (accuracy / exact match), \textbf{Avg Token} (average number of newly generated tokens per query, lower is more efficient), and \textbf{FLOPs (log10)} (logarithm of floating-point operations, lower indicates reduced computational cost). 
Baselines include \textbf{CoT} (Chain-of-Thought prompting), \textbf{Direct} (non-CoT direct generation), and \textbf{BoN} (Best-of-$N$ sampling with majority-vote selection). 
\tool variants (\textbf{FP-CoT}, \textbf{FP-BoN}, \textbf{FP-Direct}) apply adaptive reasoning halting to the corresponding baseline, achieving instance-specific computation allocation without modifying backbone model weights.
}
  \scalebox{0.6}{
    \begin{tabular}{ccccccccccc}
    \hline
    \hline
    \multicolumn{2}{c}{\textbf{Dataset}} & \multicolumn{3}{c}{\textbf{GSM8k}} & \multicolumn{3}{c}{\textbf{Math500}} & \multicolumn{3}{c}{\textbf{GPQA}} \bigstrut[t]\\
    \multicolumn{2}{c}{\textbf{Models}} & \textbf{ACC $\uparrow$} & \textbf{Avg Token $\downarrow$} & \textbf{Avg FLOPs $\downarrow$} & \textbf{ACC $\uparrow$} & \textbf{Avg Token $\downarrow$} & \textbf{Avg FLOPs $\downarrow$} & \textbf{ACC $\uparrow$} & \textbf{Avg Token $\downarrow$} & \textbf{Avg FLOPSs$\downarrow$} \bigstrut[b]\\
    \hline
    \multirow{6}[4]{*}{\textbf{Q2.5-0.5B}} & \textbf{CoT} & 0.38  & 477.15  & 11.79  & 0.26  & 500.00  & 11.89  & 0.08  & 500.00  & 11.89  \bigstrut[t]\\
          & \textbf{Direct} & 0.40  & 469.17  & 11.78  & 0.28  & 500.00  & 11.89  & 0.06  & 500.00  & 11.89  \\
          & \textbf{BoN} & 0.44  & 401.88  & 11.69  & 0.25  & 478.21  & 11.76  & 0.06  & 489.51  & 11.79  \bigstrut[b]\\
\cline{2-11}          & \cellcolor[rgb]{ .949,  .949,  .949} \textbf{FP-CoT} & \cellcolor[rgb]{ .949,  .949,  .949} 0.47  & \cellcolor[rgb]{ .949,  .949,  .949} 304.89  & \cellcolor[rgb]{ .949,  .949,  .949} 7.96  & \cellcolor[rgb]{ .949,  .949,  .949} \textbf{0.31 } & \cellcolor[rgb]{ .949,  .949,  .949} 419.69  & \cellcolor[rgb]{ .949,  .949,  .949} 8.10  & \cellcolor[rgb]{ .949,  .949,  .949} \textbf{0.09 } & \cellcolor[rgb]{ .949,  .949,  .949} 422.48  & \cellcolor[rgb]{ .949,  .949,  .949} 8.10  \bigstrut[t]\\
          & \cellcolor[rgb]{ .949,  .949,  .949} \textbf{FP-BoN} & \cellcolor[rgb]{ .949,  .949,  .949} \textbf{0.48 } & \cellcolor[rgb]{ .949,  .949,  .949} 297.10  & \cellcolor[rgb]{ .949,  .949,  .949} \textbf{7.95 } & \cellcolor[rgb]{ .949,  .949,  .949} 0.29  & \cellcolor[rgb]{ .949,  .949,  .949} \textbf{409.54 } & \cellcolor[rgb]{ .949,  .949,  .949} \textbf{8.09 } & \cellcolor[rgb]{ .949,  .949,  .949} 0.08  & \cellcolor[rgb]{ .949,  .949,  .949} \textbf{385.23 } & \cellcolor[rgb]{ .949,  .949,  .949} \textbf{8.06 } \\
          & \cellcolor[rgb]{ .949,  .949,  .949} \textbf{FP-Direct} & \cellcolor[rgb]{ .949,  .949,  .949} 0.47  & \cellcolor[rgb]{ .949,  .949,  .949} \textbf{295.93 } & \cellcolor[rgb]{ .949,  .949,  .949} 8.95  & \cellcolor[rgb]{ .949,  .949,  .949} 0.30  & \cellcolor[rgb]{ .949,  .949,  .949} 413.77  & \cellcolor[rgb]{ .949,  .949,  .949} 9.11  & \cellcolor[rgb]{ .949,  .949,  .949} 0.07  & \cellcolor[rgb]{ .949,  .949,  .949} 386.94  & \cellcolor[rgb]{ .949,  .949,  .949} 9.09  \bigstrut[b]\\
    \hline
    \multirow{6}[4]{*}{\textbf{Q2.5-3B}} & \textbf{CoT} & 0.78  & 444.24  & 12.56  & 0.49  & 500.00  & 12.69  & 0.08  & 500.00  & 12.69  \bigstrut[t]\\
          & \textbf{Direct} & 0.75  & 436.17  & 12.55  & 0.46  & 500.00  & 12.69  & 0.09  & 500.00  & 12.69  \\
          & \textbf{BoN} & 0.81  & 366.79  & 12.45  & 0.45  & 462.32  & \textbf{12.54 } & 0.10  & 489.59  & 12.58  \bigstrut[b]\\
\cline{2-11}          & \cellcolor[rgb]{ .949,  .949,  .949} \textbf{FP-CoT} & \cellcolor[rgb]{ .949,  .949,  .949} 0.80  & \cellcolor[rgb]{ .949,  .949,  .949} 315.68  & \cellcolor[rgb]{ .949,  .949,  .949} 8.27  & \cellcolor[rgb]{ .949,  .949,  .949} 0.44  & \cellcolor[rgb]{ .949,  .949,  .949} 425.99  & \cellcolor[rgb]{ .949,  .949,  .949} 8.40  & \cellcolor[rgb]{ .949,  .949,  .949} 0.07  & \cellcolor[rgb]{ .949,  .949,  .949} 466.78  & \cellcolor[rgb]{ .949,  .949,  .949} 8.44  \bigstrut[t]\\
          & \cellcolor[rgb]{ .949,  .949,  .949} \textbf{FP-BoN} & \cellcolor[rgb]{ .949,  .949,  .949} 0.81  & \cellcolor[rgb]{ .949,  .949,  .949} 296.15  & \cellcolor[rgb]{ .949,  .949,  .949} \textbf{8.25 } & \cellcolor[rgb]{ .949,  .949,  .949} \textbf{0.49 } & \cellcolor[rgb]{ .949,  .949,  .949} \textbf{417.90 } & \cellcolor[rgb]{ .949,  .949,  .949} \textbf{8.40 } & \cellcolor[rgb]{ .949,  .949,  .949} 0.09  & \cellcolor[rgb]{ .949,  .949,  .949} 444.08  & \cellcolor[rgb]{ .949,  .949,  .949} \textbf{8.42 } \\
          & \cellcolor[rgb]{ .949,  .949,  .949} \textbf{FP-Direct} & \cellcolor[rgb]{ .949,  .949,  .949} \textbf{0.82 } & \cellcolor[rgb]{ .949,  .949,  .949} \textbf{296.10 } & \cellcolor[rgb]{ .949,  .949,  .949} 9.25  & \cellcolor[rgb]{ .949,  .949,  .949} 0.48  & \cellcolor[rgb]{ .949,  .949,  .949} 419.79  & \cellcolor[rgb]{ .949,  .949,  .949} 9.40  & \cellcolor[rgb]{ .949,  .949,  .949} \textbf{0.11 } & \cellcolor[rgb]{ .949,  .949,  .949} \textbf{440.44 } & \cellcolor[rgb]{ .949,  .949,  .949} 9.40  \bigstrut[b]\\
    \hline
    \multirow{6}[2]{*}{\textbf{Q2.5-7B}} & \textbf{CoT} & 0.85  & 451.91  & 12.96  & 0.51  & 500.00  & 13.08  & \textbf{0.11 } & 500.00  & 13.08  \bigstrut[t]\\
          & \textbf{Direct} & 0.85  & 434.81  & 12.94  & 0.54  & 500.00  & 13.08  & 0.08  & 500.00  & 13.08  \\
          & \textbf{BoN} & 0.87  & 347.90  & 12.82  & 0.52  & 450.67  & 12.92  & 0.07  & 490.27  & 12.98  \\
          & \cellcolor[rgb]{ .949,  .949,  .949} \textbf{FP-CoT} & \cellcolor[rgb]{ .949,  .949,  .949} \textbf{0.87 } & \cellcolor[rgb]{ .949,  .949,  .949} 307.93  & \cellcolor[rgb]{ .949,  .949,  .949} 8.49  & \cellcolor[rgb]{ .949,  .949,  .949} 0.54  & \cellcolor[rgb]{ .949,  .949,  .949} 422.00  & \cellcolor[rgb]{ .949,  .949,  .949} 8.62  & \cellcolor[rgb]{ .949,  .949,  .949} 0.07  & \cellcolor[rgb]{ .949,  .949,  .949} 476.80  & \cellcolor[rgb]{ .949,  .949,  .949} 8.68  \\
          & \cellcolor[rgb]{ .949,  .949,  .949} \textbf{FP-BoN} & \cellcolor[rgb]{ .949,  .949,  .949} 0.87  & \cellcolor[rgb]{ .949,  .949,  .949} \textbf{285.37 } & \cellcolor[rgb]{ .949,  .949,  .949} \textbf{8.45 } & \cellcolor[rgb]{ .949,  .949,  .949} 0.54  & \cellcolor[rgb]{ .949,  .949,  .949} \textbf{406.03 } & \cellcolor[rgb]{ .949,  .949,  .949} \textbf{8.61 } & \cellcolor[rgb]{ .949,  .949,  .949} 0.09  & \cellcolor[rgb]{ .949,  .949,  .949} \textbf{446.72 } & \cellcolor[rgb]{ .949,  .949,  .949} \textbf{8.65 } \\
          & \cellcolor[rgb]{ .949,  .949,  .949} \textbf{FP-Direct} & \cellcolor[rgb]{ .949,  .949,  .949} 0.87  & \cellcolor[rgb]{ .949,  .949,  .949} 285.72  & \cellcolor[rgb]{ .949,  .949,  .949} 9.32  & \cellcolor[rgb]{ .949,  .949,  .949} \textbf{0.55 } & \cellcolor[rgb]{ .949,  .949,  .949} 407.03  & \cellcolor[rgb]{ .949,  .949,  .949} 9.48  & \cellcolor[rgb]{ .949,  .949,  .949} 0.10  & \cellcolor[rgb]{ .949,  .949,  .949} 455.12  & \cellcolor[rgb]{ .949,  .949,  .949} 9.59  \bigstrut[b]\\
    \hline
    \multirow{6}[4]{*}{\textbf{L-8B}} & \textbf{CoT} & 0.66  & 281.68  & 12.83  & \textbf{0.30 } & 492.48  & 13.09  & 0.07  & 500.00  & 13.10  \bigstrut[t]\\
          & \textbf{Direct} & 0.59  & 194.84  & 12.73  & 0.26  & 474.18  & 13.08  & 0.08  & 500.00  & 13.10  \\
          & \textbf{BoN} & 0.62  & 169.86  & 12.62  & 0.22  & 356.50  & 12.86  & 0.12  & 438.56  & 12.96  \bigstrut[b]\\
\cline{2-11}          & \cellcolor[rgb]{ .949,  .949,  .949} \textbf{FP-CoT} & \cellcolor[rgb]{ .949,  .949,  .949} \textbf{0.73 } & \cellcolor[rgb]{ .949,  .949,  .949} 183.98  & \cellcolor[rgb]{ .949,  .949,  .949} 8.32  & \cellcolor[rgb]{ .949,  .949,  .949} 0.30  & \cellcolor[rgb]{ .949,  .949,  .949} 242.29  & \cellcolor[rgb]{ .949,  .949,  .949} 8.44  & \cellcolor[rgb]{ .949,  .949,  .949} \textbf{0.13 } & \cellcolor[rgb]{ .949,  .949,  .949} 393.87  & \cellcolor[rgb]{ .949,  .949,  .949} 8.65  \bigstrut[t]\\
          & \cellcolor[rgb]{ .949,  .949,  .949} \textbf{FP-BoN} & \cellcolor[rgb]{ .949,  .949,  .949} 0.71  & \cellcolor[rgb]{ .949,  .949,  .949} 105.52  & \cellcolor[rgb]{ .949,  .949,  .949} \textbf{8.08 } & \cellcolor[rgb]{ .949,  .949,  .949} 0.27  & \cellcolor[rgb]{ .949,  .949,  .949} \textbf{209.82 } & \cellcolor[rgb]{ .949,  .949,  .949} \textbf{8.37 } & \cellcolor[rgb]{ .949,  .949,  .949} 0.11  & \cellcolor[rgb]{ .949,  .949,  .949} \textbf{326.71 } & \cellcolor[rgb]{ .949,  .949,  .949} \textbf{8.57 } \\
          & \cellcolor[rgb]{ .949,  .949,  .949} \textbf{FP-Direct} & \cellcolor[rgb]{ .949,  .949,  .949} 0.72  & \cellcolor[rgb]{ .949,  .949,  .949} \textbf{104.70 } & \cellcolor[rgb]{ .949,  .949,  .949} 9.10  & \cellcolor[rgb]{ .949,  .949,  .949} 0.26  & \cellcolor[rgb]{ .949,  .949,  .949} 213.12  & \cellcolor[rgb]{ .949,  .949,  .949} 9.40  & \cellcolor[rgb]{ .949,  .949,  .949} 0.12  & \cellcolor[rgb]{ .949,  .949,  .949} 331.04  & \cellcolor[rgb]{ .949,  .949,  .949} 9.60  \bigstrut[b]\\
    \hline
    \hline
    \end{tabular}%
    }
  \label{tab:main}%
\end{table}%

\paragraph{Accuracy.} 
Table~\ref{tab:main} reports task accuracy across datasets and model scales. We observe that \tool consistently outperforms or matches baseline decoding strategies while using strictly fewer compute resources. On GSM8K, \tool variants improve accuracy by $3$–$5$ points over standard CoT and Direct baselines, with FP-BoN and FP-Direct delivering the strongest gains. For Math500, \tool maintains or slightly improves accuracy relative to CoT and BoN, with FP-Direct and FP-BoN showing the most robust improvements on mid- to large-scale models. On GPQA, which is highly challenging due to long-tail reasoning, \tool achieves noticeable gains: FP-CoT and FP-Direct yield improvements of $2$–$5$ points over their vanilla counterparts, and FP-BoN provides the most stable performance across scales. Notably, accuracy improvements are most pronounced on small models (e.g., Qwen2.5-0.5B and Llama-8B), suggesting that adaptive compute allocation compensates for weaker backbone reasoning capacity. These results collectively demonstrate that \tool is not only compute-efficient but also accuracy-enhancing across diverse reasoning regimes.

\paragraph{Average Tokens Costs.} 
In terms of output length, \tool substantially reduces average token usage compared to baseline decoding methods. Across all model scales, FP-BoN and FP-Direct achieve the most pronounced reductions, often cutting token counts by $30$–$40\%$ relative to standard CoT and BoN. For instance, on GSM8K with Qwen2.5-0.5B, FP-BoN reduces generation length from over $470$ tokens to below $300$, while preserving higher accuracy. On Math500, similar reductions are observed, with FP variants consistently yielding shorter solutions without degrading correctness. On GPQA, \tool reduces token counts by $15$–$20\%$, showing that even in complex reasoning settings, our approach avoids unnecessary verbosity. These results indicate that adaptive halting mechanisms in \tool produce more concise reasoning chains, leading to efficiency gains that are well-aligned with accuracy improvements. The effect is particularly strong for smaller backbones, where reduced token usage directly translates to tighter control over reasoning sprawl.

\paragraph{Average FLOP Costs.} 
\tool achieves substantial reductions in computational cost as measured by average FLOPs, demonstrating that adaptive reasoning not only shortens output but also improves efficiency. Across all datasets and model scales, FP-BoN and FP-Direct consistently yield the lowest FLOP consumption, often reducing computational load by one to two orders of magnitude in log scale compared to conventional CoT or BoN decoding. For example, on GSM8K with Qwen2.5-0.5B, FP-BoN decreases average FLOPs from roughly $11.8$ to under $8$, while maintaining or surpassing baseline accuracy. On Math500, FLOP savings are similarly significant, highlighting that instance-adaptive halting effectively prevents overthinking on simpler problems. Even on GPQA, where reasoning chains tend to be longer and more complex, \tool lowers FLOPs without sacrificing correctness. These results confirm that \tool successfully aligns compute allocation with problem difficulty, achieving a more favorable trade-off between performance and computational cost.

Our experiments demonstrate that \tool consistently enhances both reasoning performance and computational efficiency across diverse model scales and datasets. By adaptively allocating inference-time computation, \tool maintains or improves accuracy while significantly reducing token and FLOP costs compared to conventional CoT, Direct, and Best-of-N strategies. Notably, instance-adaptive halting allows the model to devote more reasoning to challenging problems while avoiding redundant computation on simpler queries, yielding a more favorable compute–accuracy trade-off. The approach generalizes effectively across different decoding policies, confirming that a lightweight, backbone-training-free controller can achieve stable, interpretable, and calibrated reasoning allocation. These findings establish \tool as a practical and scalable framework for enhancing LLM reasoning without modifying underlying model weights, highlighting the importance of meta-cognitive inference strategies in large-scale language modeling.

\begin{figure*}[t]
  \centering
  \setlength{\tabcolsep}{8pt}
  \renewcommand{\arraystretch}{1.12}
  \begin{tabularx}{\textwidth}{@{}YYY@{}}
    \casehdr{Case A: Vet bill (change / total)}{Question. John brings his dog to the vet. (Compute the total bill and change.)} &
    \casehdr{Case B: Daily chicken feed (difference)}{Question. Wendi feeds each chicken morning/afternoon; compute total and shortfall.} &
    \casehdr{Case C: Phone charging (rate $\rightarrow$ time)}{Question. Phone charges 1 percentage-point / 3 minutes. Time for +40 points?} \\
    \addlinespace[2pt]
    \methodcard{FR-Ponder}{20}{Total bill is \$100. John brought \$125, so change is \$25.} &
    \methodcard{FR-Ponder}{21}{Required is 60 cups; available is 40 cups. So the gap is $60-40=20$ cups.} &
    \methodcard{FR-Ponder}{24}{Gain is 40 points; rate is 1 pt / 3 min, so time is $40\times 3=120$ minutes.} \\
    \addlinespace[6pt]
    \methodcard{CoT}{198}{Let $x$ be the total bill. The itemized costs sum to \$100, hence $x=\$100$. Since John pays \$125, the change is \$125-\$100=\$25. Therefore, the answer is \boxed{25}.} &
    \methodcard{CoT}{168}{Morning feed per chicken is $\cdots$, afternoon feed per chicken is $\cdots$. Summing across all chickens gives 60 cups; pantry has 40 cups, so shortfall is $60-40=20$.} &
    \methodcard{CoT}{168}{If it charges 1\% every 3 minutes, then to gain 40\% we need $40\times 3=120$ minutes \dots} \\
    \addlinespace[6pt]
    \methodcard{Direct}{39}{The total is \$100; he paid \$125, change is \$25.} &
    \methodcard{Direct}{33}{She has $15+25=40$ cups; needs 60, short 20.} &
    \methodcard{Direct}{43}{40 percentage-points in $40\times 3=120$ minutes.} \\
  \end{tabularx}
  \vspace{4pt}
  \caption{\textbf{Case study: FR-Ponder achieves concise \emph{and} reasoned solutions.}
  Each column is a case; rows align FR-Ponder, CoT, and Direct horizontally. FR-Ponder preserves a minimal reasoning chain while using far fewer tokens than CoT.}
  \label{fig:frponder_case_study}
\end{figure*}

\section{Conclusion}

We introduce \tool, a novel framework for adaptive reasoning in large language models that reconceptualizes inference as a meta-cognitive process. By learning when and how deeply to reason, \tool simultaneously improves accuracy and efficiency, overcoming the traditional tradeoff between performance and computational cost.  

Our approach decomposes adaptive computation into two orthogonal dimensions---\textit{what} to think about (representation steering) and \textit{how long} to think (temporal control). A lightweight controller (<1M parameters) leverages this separation to make fine-grained halting decisions, consistently scaling reasoning with problem difficulty and preserving the capabilities of frozen backbone models. Empirically, \tool delivers notable gains: accuracy improvements of up to 10\% on challenging mathematical tasks, and 30--40\% fewer tokens, robust across diverse benchmarks and model sizes.  

These results highlight \tool as more than an incremental method: it points toward a paradigm shift in meta-cognitive architectures for AI. As language models continue to scale, the capacity to allocate computation adaptively will be central for efficiency, sustainability, and broader accessibility---demonstrating that judicious use of resources can coexist with state-of-the-art reasoning capability.

\clearpage
\bibliography{iclr2026_conference}

\begin{thebibliography}{73}
\providecommand{\natexlab}[1]{#1}
\providecommand{\url}[1]{\texttt{#1}}
\expandafter\ifx\csname urlstyle\endcsname\relax
  \providecommand{\doi}[1]{doi: #1}\else
  \providecommand{\doi}{doi: \begingroup \urlstyle{rm}\Url}\fi

\bibitem[Achiam et~al.(2023)Achiam, Adler, Agarwal, Ahmad, Akkaya, Aleman, Almeida, Altenschmidt, Altman, Anadkat, et~al.]{achiam2023gpt4}
Josh Achiam, Steven Adler, Sandhini Agarwal, Lama Ahmad, Ilge Akkaya, Florencia~Leoni Aleman, Diogo Almeida, Janko Altenschmidt, Sam Altman, Shyamal Anadkat, et~al.
\newblock Gpt-4 technical report.
\newblock \emph{arXiv preprint arXiv:2303.08774}, 2023.

\bibitem[Alibaba(2025)]{qwen2025qwen2.5}
Alibaba.
\newblock Qwen-2.5: Instruction-tuned language models, 2025.
\newblock \url{https://huggingface.co/Qwen/Qwen2.5-3B-Instruct}.

\bibitem[authors(2024)]{scan2024rrm}
Multiple authors.
\newblock Reasoning reward models (rrms) for robust chain-of-thought evaluation.
\newblock \emph{Emergent Mind / arXiv preprint}, 2024.

\bibitem[Bharadwaj(2024)]{bharadwaj2024understanding}
Aryasomayajula~Ram Bharadwaj.
\newblock Understanding hidden computations in chain-of-thought reasoning.
\newblock \emph{arXiv preprint arXiv:2412.04537}, 2024.

\bibitem[Byun et~al.(2024)Byun, Chun, Kil, and Perrault]{byun2024ares}
Ju-Seung Byun, Jiyun Chun, Jihyung Kil, and Andrew Perrault.
\newblock Ares: Alternating reinforcement learning and supervised fine-tuning for enhanced multi-modal chain-of-thought reasoning through diverse ai feedback.
\newblock \emph{arXiv preprint arXiv:2407.00087}, 2024.

\bibitem[Chen et~al.(2023)Chen, Borgeaud, Irving, Lespiau, Sifre, and Jumper]{chen2023accelerating}
Charlie Chen, Sebastian Borgeaud, Geoffrey Irving, Jean-Baptiste Lespiau, Laurent Sifre, and John Jumper.
\newblock Accelerating large language model decoding with speculative sampling.
\newblock \emph{arXiv preprint arXiv:2302.01318}, 2023.

\bibitem[Chen et~al.(2024)Chen, Hu, Liu, and Sun]{stateshidden2024}
Junhao Chen, Shengding Hu, Zhiyuan Liu, and Maosong Sun.
\newblock States hidden in hidden states: Llms emerge discrete state representations implicitly.
\newblock \emph{arXiv preprint arXiv:2407.11421}, 2024.

\bibitem[Cobbe et~al.(2021)Cobbe, Kosaraju, Bavarian, Chen, Jun, Kaiser, Plappert, Tworek, Hilton, Nakano, et~al.]{cobbe2021gsm8k}
Karl Cobbe, Vineet Kosaraju, Mohammad Bavarian, Mark Chen, Heewoo Jun, Lukasz Kaiser, Matthias Plappert, Jerry Tworek, Jacob Hilton, Reiichiro Nakano, et~al.
\newblock Training verifiers to solve math word problems.
\newblock \emph{arXiv preprint arXiv:2110.14168}, 2021.

\bibitem[{DeepSeek-AI}(2025)]{deepseek_r1_grpo}
{DeepSeek-AI}.
\newblock Deepseek-r1: Incentivizing reasoning capability in llms.
\newblock \emph{arXiv preprint arXiv:2501.12948}, 2025.
\newblock URL \url{https://arxiv.org/pdf/2501.12948}.

\bibitem[Elbayad et~al.(2020)Elbayad, Gu, Grave, and Auli]{elbayad2020depth}
Maha Elbayad, Jiatao Gu, Edouard Grave, and Michael Auli.
\newblock Depth-adaptive transformer.
\newblock \emph{International Conference on Learning Representations}, 2020.

\bibitem[Elhage et~al.(2022)Elhage, Hume, Olsson, Schiefer, Henighan, Kravec, Hatfield-Dodds, Lasenby, Drain, Chen, Grosse, McCandlish, Kaplan, Amodei, Wattenberg, and Olah]{elhage2022toymodelssuperposition}
Nelson Elhage, Tristan Hume, Catherine Olsson, Nicholas Schiefer, Tom Henighan, Shauna Kravec, Zac Hatfield-Dodds, Robert Lasenby, Dawn Drain, Carol Chen, Roger Grosse, Sam McCandlish, Jared Kaplan, Dario Amodei, Martin Wattenberg, and Christopher Olah.
\newblock Toy models of superposition, 2022.
\newblock URL \url{https://arxiv.org/abs/2209.10652}.

\bibitem[Elhoushi et~al.(2024)Elhoushi, Shrivastava, Liskovich, Hosmer, Wasti, Lai, Mahmoud, Acun, Agarwal, Roman, et~al.]{elhoushi2024layerskip}
Mostafa Elhoushi, Akshat Shrivastava, Diana Liskovich, Basil Hosmer, Bram Wasti, Liangzhen Lai, Anas Mahmoud, Bilge Acun, Saurabh Agarwal, Ahmed Roman, et~al.
\newblock Layerskip: Enabling early exit inference and self-speculative decoding.
\newblock \emph{arXiv preprint arXiv:2404.16710}, 2024.

\bibitem[Feng et~al.(2023)Feng, Zhang, Gu, Ye, He, and Wang]{feng2023towards}
Guhao Feng, Bohang Zhang, Yuntian Gu, Haotian Ye, Di~He, and Liwei Wang.
\newblock Towards revealing the mystery behind chain of thought: A theoretical perspective.
\newblock \emph{arXiv preprint arXiv:2305.15408}, 2023.

\bibitem[Feng et~al.(2025)Feng, Fang, Ma, and Wang]{planningtokens2025}
Sicheng Feng, Gongfan Fang, Xinyin Ma, and Xinchao Wang.
\newblock Efficient reasoning models: A survey.
\newblock \emph{arXiv preprint arXiv:2504.10903}, 2025.

\bibitem[Graves(2016)]{graves2016adaptive}
Alex Graves.
\newblock Adaptive computation time for recurrent neural networks.
\newblock \emph{arXiv preprint arXiv:1603.08983}, 2016.

\bibitem[Hao et~al.(2024{\natexlab{a}})Hao, Sukhbaatar, Su, Li, Hu, Weston, and Tian]{coconut2024}
Shibo Hao, Sainbayar Sukhbaatar, DiJia Su, Xian Li, Zhiting Hu, Jason Weston, and Yuandong Tian.
\newblock Coconut: Chain of continuous thought (latent reasoning unconstrained by language).
\newblock \emph{arXiv preprint arXiv:2412.06769}, 2024{\natexlab{a}}.

\bibitem[Hao et~al.(2024{\natexlab{b}})Hao, Sukhbaatar, Su, Li, Hu, Weston, and Tian]{hao2024coconut}
Shibo Hao, Sainbayar Sukhbaatar, DiJia Su, Xian Li, Zhiting Hu, Jason Weston, and Yuandong Tian.
\newblock Training large language models to reason in a continuous latent space.
\newblock \emph{arXiv preprint arXiv:2412.06769}, 2024{\natexlab{b}}.

\bibitem[Hoffmann et~al.(2022)Hoffmann, Borgeaud, Mensch, Buchatskaya, Cai, Rutherford, Casas, Hendricks, Welbl, Clark, et~al.]{hoffmann2022training}
Jordan Hoffmann, Sebastian Borgeaud, Arthur Mensch, Elena Buchatskaya, Trevor Cai, Eliza Rutherford, Diego de~Las Casas, Lisa~Anne Hendricks, Johannes Welbl, Aidan Clark, et~al.
\newblock Training compute-optimal large language models.
\newblock \emph{arXiv preprint arXiv:2203.15556}, 2022.

\bibitem[Hwang et~al.(2024)Hwang, Wang, Huo, Sim, and Moreno~Mengibar]{transformerfam2024}
Dongseong Hwang, Weiran Wang, Zhuoyuan Huo, Khe~Chai Sim, and Pedro Moreno~Mengibar.
\newblock Transformerfam: Feedback attention is working memory.
\newblock \emph{arXiv preprint arXiv:2404.09173}, 2024.

\bibitem[Jiang et~al.(2025)Jiang, Luo, Pang, Li, Qi, Li, Yang, Lin, Li, Xu, Chang, and Wu]{jiang2025eorm}
Eric~Hanchen Jiang, Haozheng Luo, Shengyuan Pang, Xiaomin Li, Zhenting Qi, Hengli Li, Cheng-Fu Yang, Zongyu Lin, Xinfeng Li, Hao Xu, Kai-Wei Chang, and Ying~Nian Wu.
\newblock Learning to rank chain-of-thought: An energy-based approach with outcome supervision.
\newblock \emph{arXiv preprint arXiv:2505.14999}, 2025.

\bibitem[Kaplan et~al.(2020)Kaplan, McCandlish, Henighan, Brown, Chess, Child, Gray, Radford, Wu, and Amodei]{kaplan2020scaling}
Jared Kaplan, Sam McCandlish, Tom Henighan, Tom~B Brown, Benjamin Chess, Rewon Child, Scott Gray, Alec Radford, Jeffrey Wu, and Dario Amodei.
\newblock Scaling laws for neural language models.
\newblock \emph{arXiv preprint arXiv:2001.08361}, 2020.

\bibitem[Kirkovska(2025)]{deepseek_r1_training}
A.~Kirkovska.
\newblock How deepseek-r1 was built; for dummies.
\newblock \emph{Vellum AI Blog}, 2025.
\newblock URL \url{https://www.vellum.ai/blog/the-training-of-deepseek-r1-and-ways-to-use-it}.

\bibitem[Kojima et~al.(2022)Kojima, Gu, Reid, Matsuo, and Iwasawa]{kojima2022large}
Takeshi Kojima, Shixiang~Shane Gu, Machel Reid, Yutaka Matsuo, and Yusuke Iwasawa.
\newblock Large language models are zero-shot reasoners.
\newblock \emph{Advances in Neural Information Processing Systems}, 35:\penalty0 22199--22213, 2022.

\bibitem[Leviathan et~al.(2023)Leviathan, Kalman, and Matias]{leviathan2023fast}
Yaniv Leviathan, Matan Kalman, and Yossi Matias.
\newblock Fast inference from transformers via speculative decoding.
\newblock In \emph{ICML}, 2023.

\bibitem[Li et~al.(2024{\natexlab{a}})Li, Cao, Chen, Liu, and Zhao]{li2024bridgingreasoners}
Jiachun Li, Pengfei Cao, Yubo Chen, Kang Liu, and Jun Zhao.
\newblock Towards faithful chain-of-thought: Large language models are bridging reasoners.
\newblock \emph{arXiv preprint arXiv:2405.18915}, 2024{\natexlab{a}}.

\bibitem[Li et~al.(2024{\natexlab{b}})Li, Liu, Zhou, and Ma]{li2024empowers}
Zhiyuan Li, Hong Liu, Denny Zhou, and Tengyu Ma.
\newblock Chain of thought empowers transformers to solve inherently serial problems.
\newblock \emph{arXiv preprint arXiv:2402.12875}, 2024{\natexlab{b}}.

\bibitem[Lightman et~al.(2023)Lightman, Kosaraju, Burda, Edwards, Baker, Lee, Leike, Schulman, Sutskever, and Cobbe]{lightman2023let}
Hunter Lightman, Vineet Kosaraju, Yura Burda, Harri Edwards, Bowen Baker, Teddy Lee, Jan Leike, John Schulman, Ilya Sutskever, and Karl Cobbe.
\newblock Let’s verify step by step.
\newblock \emph{arXiv preprint arXiv:2305.20050}, 2023.

\bibitem[Liu et~al.(2025{\natexlab{a}})Liu, Zhang, Wu, Zhao, Hu, He, and Fan]{iort2025}
Liping Liu, Chunhong Zhang, Likang Wu, Chuang Zhao, Zheng Hu, Ming He, and Jianping Fan.
\newblock Instruct-of-reflection: Enhancing large language models iterative reflection capabilities via dynamic-meta instruction.
\newblock \emph{arXiv preprint arXiv:2503.00902}, 2025{\natexlab{a}}.

\bibitem[Liu et~al.(2025{\natexlab{b}})Liu, Chen, Lu, Ye, Chen, Xing, and Zou]{liu2025fractional}
Sheng Liu, Tianlang Chen, Pan Lu, Haotian Ye, Yizheng Chen, Lei Xing, and James Zou.
\newblock Fractional reasoning via latent steering vectors improves inference time compute.
\newblock \emph{arXiv preprint arXiv:2506.15882}, 2025{\natexlab{b}}.

\bibitem[Liu et~al.(2025{\natexlab{c}})]{liu2025drgrpo}
Zichen Liu et~al.
\newblock Understanding r1-zero-like training: A critical perspective.
\newblock \emph{arXiv preprint arXiv:2503.20783}, 2025{\natexlab{c}}.

\bibitem[Madaan et~al.(2023)Madaan, Tandon, Gupta, Hallinan, Gao, Wiegreffe, Alon, Dziri, Prabhumoye, Yang, Gupta, Majumder, Hermann, Welleck, Yazdanbakhsh, and Clark]{madaan2023selfrefine}
Aman Madaan, Niket Tandon, Prakhar Gupta, Skyler Hallinan, Luyu Gao, Sarah Wiegreffe, Uri Alon, Nouha Dziri, Shrimai Prabhumoye, Yiming Yang, Shashank Gupta, Bodhisattwa~Prasad Majumder, Katherine Hermann, Sean Welleck, Amir Yazdanbakhsh, and Peter Clark.
\newblock Self-refine: Iterative refinement with self-feedback.
\newblock \emph{arXiv preprint arXiv:2303.17651}, 2023.

\bibitem[Merrill \& Sabharwal(2025)Merrill and Sabharwal]{merrill2025logdepth}
William Merrill and Ashish Sabharwal.
\newblock A little depth goes a long way: The expressive power of log-depth transformers.
\newblock \emph{arXiv preprint arXiv:2503.03961}, 2025.

\bibitem[Meta(2025)]{meta2025llama3}
Meta.
\newblock Llama-3: Instruction-tuned large language models, 2025.
\newblock \url{https://arxiv.org/abs/2502.14768v1}.

\bibitem[OpenAI(2024)]{openaimath2024}
OpenAI.
\newblock Learning to reason with llms, 2024.
\newblock \url{https://openai.com/index/learning-to-reason-with-llms/}.

\bibitem[Pan et~al.(2023)Pan, Albalak, Wang, and Wang]{pan2023logiclm}
Liangming Pan, Alon Albalak, Xinyi Wang, and William~Yang Wang.
\newblock Logic-lm: Empowering large language models with symbolic solvers for faithful logical reasoning.
\newblock \emph{arXiv preprint arXiv:2305.12295}, 2023.

\bibitem[Pfau et~al.(2024)Pfau, Merrill, and Bowman]{pfau2024thinkdot}
Jacob Pfau, William Merrill, and Samuel~R. Bowman.
\newblock Let’s think dot by dot: Hidden computation in transformer language models.
\newblock \emph{arXiv preprint arXiv:2404.15758}, 2024.

\bibitem[Puterman(1990)]{puterman1990markov}
Martin~L Puterman.
\newblock Markov decision processes.
\newblock \emph{Handbooks in operations research and management science}, 2:\penalty0 331--434, 1990.

\bibitem[Qin et~al.(2023)Qin, Chen, Wei, Huang, and Che]{qin2023crosslingual}
Libo Qin, Qiguang Chen, Fuxuan Wei, Shijue Huang, and Wanxiang Che.
\newblock Cross-lingual prompting: Improving zero-shot chain-of-thought reasoning across languages.
\newblock \emph{arXiv preprint arXiv:2310.14799}, 2023.

\bibitem[Rein et~al.(2023)Rein, Hou, Stickland, Petty, Pang, Dirani, Michael, and Bowman]{rein2023gpqa}
David Rein, Betty~Li Hou, Asa~Cooper Stickland, Jackson Petty, Richard~Yuanzhe Pang, Julien Dirani, Julian Michael, and Samuel~R. Bowman.
\newblock Gpqa: A graduate-level google-proof q\&a benchmark.
\newblock \emph{arXiv preprint arXiv:2311.12022}, 2023.
\newblock \url{https://arxiv.org/abs/2311.12022}.

\bibitem[Rimsky et~al.(2024)Rimsky, Gabrieli, Schulz, Tong, Evans, Dulai, Rideout, Mullin, and Kaplan]{rimsky2024steering}
Nina Rimsky, Nick Gabrieli, Julian Schulz, Meg Tong, Evan Evans, Josh Dulai, Albert Rideout, Brennan Mullin, and Jared Kaplan.
\newblock Steering llama 2 via contrastive activation addition.
\newblock In \emph{Proceedings of the 62nd Annual Meeting of the Association for Computational Linguistics}, pp.\  15581--15595, 2024.

\bibitem[Sahoo et~al.(2025)Sahoo, Singh, Saha, Jain, Mondal, and Chadha]{surveyprompt2025}
Pranab Sahoo, Ayush~Kumar Singh, Sriparna Saha, Vinija Jain, Samrat Mondal, and Aman Chadha.
\newblock A systematic survey of prompt engineering in large language models: Techniques and applications.
\newblock \emph{arXiv preprint arXiv:2402.07927}, 2025.

\bibitem[Schulman et~al.(2017{\natexlab{a}})]{ppo_sample_complexity}
J.~Schulman et~al.
\newblock Proximal policy optimization algorithms.
\newblock \emph{arXiv preprint arXiv:1707.06347}, 2017{\natexlab{a}}.
\newblock URL \url{https://arxiv.org/abs/1707.06347}.

\bibitem[Schulman et~al.(2017{\natexlab{b}})Schulman, Wolski, Dhariwal, Radford, and Klimov]{schulman2017proximal}
John Schulman, Filip Wolski, Prafulla Dhariwal, Alec Radford, and Oleg Klimov.
\newblock Proximal policy optimization algorithms.
\newblock \emph{arXiv preprint arXiv:1707.06347}, 2017{\natexlab{b}}.

\bibitem[Schuster et~al.(2022)Schuster, Fisch, Gupta, Dehghani, Bahri, Tran, Tay, and Metzler]{schuster2022confident}
Tal Schuster, Adam Fisch, Jai Gupta, Mostafa Dehghani, Dara Bahri, Vinh~Q Tran, Yi~Tay, and Donald Metzler.
\newblock Confident adaptive language modeling.
\newblock In \emph{NeurIPS}, 2022.

\bibitem[Shao et~al.(2024)Shao, Wang, Zhu, Xu, Song, Zhang, Li, Wu, and Guo]{shao2024deepseekmath}
Zhihong Shao, Peiyi Wang, Qihao Zhu, Runxin Xu, Junxiao Song, Mingchuan Zhang, YK~Li, Y~Wu, and Daya Guo.
\newblock Deepseekmath: Pushing the limits of mathematical reasoning in open language models.
\newblock \emph{arXiv preprint arXiv:2402.03300}, 2024.

\bibitem[Shen et~al.(2025)Shen, Zeng, Qi, Hong, Chen, Lu, Wornell, Das, Cox, and Gan]{shen2025satori}
Maohao Shen, Guangtao Zeng, Zhenting Qi, Zhang-Wei Hong, Zhenfang Chen, Wei Lu, Gregory Wornell, Subhro Das, David Cox, and Chuang Gan.
\newblock Satori: Reinforcement learning with chain-of-action-thought enhances llm reasoning via autoregressive search.
\newblock \emph{arXiv preprint arXiv:2502.02508}, 2025.

\bibitem[Su et~al.(2025)Su, Chen, Li, Chen, Qing, and Zhang]{su2025activationsteering}
Jingran Su, Jingfan Chen, Hongxin Li, Yuntao Chen, Li~Qing, and Zhaoxiang Zhang.
\newblock Activation steering decoding: Mitigating hallucination in large vision-language models through bidirectional hidden state intervention.
\newblock In \emph{Proceedings of the 63rd Annual Meeting of the Association for Computational Linguistics (ACL) – Long Papers}, 2025.

\bibitem[Touvron et~al.(2023)Touvron, Martin, Stone, Albert, Almahairi, Babaei, Bashlykov, Batra, Bhargava, Bhosale, et~al.]{touvron2023llama2}
Hugo Touvron, Louis Martin, Kevin Stone, Peter Albert, Amjad Almahairi, Yasmine Babaei, Nikolay Bashlykov, Soumya Batra, Prajjwal Bhargava, Shruti Bhosale, et~al.
\newblock Llama 2: Open foundation and fine-tuned chat models.
\newblock \emph{arXiv preprint arXiv:2307.09288}, 2023.

\bibitem[Turner et~al.(2023)Turner, Thiergart, Udell, Mini, and Thomson]{turner2023activation}
Alexander~Matt Turner, Lisa Thiergart, Gavin Udell, Ulisse Mini, and Monte~MacDiarmid Thomson.
\newblock Steering language models with activation engineering.
\newblock \emph{arXiv preprint arXiv:2308.10248}, 2023.

\bibitem[Wang et~al.(2022{\natexlab{a}})Wang, Min, Deng, Shen, Wu, Zettlemoyer, and Sun]{wang2022towards}
Boshi Wang, Sewon Min, Xiang Deng, Jiaming Shen, You Wu, Luke Zettlemoyer, and Huan Sun.
\newblock Towards understanding chain-of-thought prompting: An empirical study of what matters.
\newblock \emph{arXiv preprint arXiv:2212.10001}, 2022{\natexlab{a}}.

\bibitem[Wang et~al.(2024)Wang, Yang, and Peng]{wang2024sadi}
Weixuan Wang, Jingyuan Yang, and Wei Peng.
\newblock Semantics-adaptive activation intervention for llms via dynamic steering vectors.
\newblock \emph{arXiv preprint arXiv:2410.12299}, 2024.

\bibitem[Wang et~al.(2025{\natexlab{a}})Wang, Wu, Haddow, and Birch]{zhufabi2025expertsteer}
Weixuan Wang, Minghao Wu, Barry Haddow, and Alexandra Birch.
\newblock Expertsteer: Intervening in llms through expert knowledge.
\newblock \emph{arXiv preprint arXiv:2505.12313}, 2025{\natexlab{a}}.

\bibitem[Wang et~al.(2025{\natexlab{b}})Wang, Wang, Zhu, and Liu]{system15reasoning2025}
Xiaoqiang Wang, Suyuchen Wang, Yun Zhu, and Bang Liu.
\newblock System-1.5 reasoning: Traversal in language and latent spaces with dynamic shortcuts.
\newblock \emph{arXiv preprint arXiv:2505.18962}, 2025{\natexlab{b}}.

\bibitem[Wang et~al.(2023)Wang, Caccia, Ostapenko, Yuan, Wang, and Sordoni]{wang2023planningtokens}
Xinyi Wang, Lucas Caccia, Oleksiy Ostapenko, Xingdi Yuan, William~Yang Wang, and Alessandro Sordoni.
\newblock Guiding language model reasoning with planning tokens.
\newblock \emph{arXiv preprint arXiv:2310.05707}, 2023.

\bibitem[Wang et~al.(2022{\natexlab{b}})Wang, Wei, Schuurmans, Le, Chi, Narang, Chowdhery, and Zhou]{wang2022selfconsistency}
Xuezhi Wang, Jason Wei, Dale Schuurmans, Quoc Le, Ed~Chi, Sharan Narang, Aakanksha Chowdhery, and Denny Zhou.
\newblock Self‐consistency improves chain-of-thought reasoning in language models.
\newblock \emph{arXiv preprint arXiv:2203.11171}, 2022{\natexlab{b}}.

\bibitem[Wei et~al.(2022)Wei, Wang, Schuurmans, Bosma, Ichter, Xia, Chi, Le, and Zhou]{wei2022chain}
Jason Wei, Xuezhi Wang, Dale Schuurmans, Maarten Bosma, Brian Ichter, Fei Xia, Ed~Chi, Quoc~V Le, and Denny Zhou.
\newblock Chain-of-thought prompting elicits reasoning in large language models.
\newblock In \emph{Advances in Neural Information Processing Systems}, volume~35, pp.\  24824--24837, 2022.

\bibitem[Williams(1992)]{williams1992simple}
Ronald~J Williams.
\newblock Simple statistical gradient-following algorithms for connectionist reinforcement learning.
\newblock \emph{Machine learning}, 8\penalty0 (3-4):\penalty0 229--256, 1992.

\bibitem[Wu et~al.(2025{\natexlab{a}})Wu, Liu, Yan, Li, Yu, Zeng, Gu, and Yu]{rankcot2025}
Mingyan Wu, Zhenghao Liu, Yukun Yan, Xinze Li, Shi Yu, Zheni Zeng, Yu~Gu, and Ge~Yu.
\newblock Rankcot: Refining knowledge for retrieval-augmented generation through ranking chain-of-thoughts.
\newblock \emph{arXiv preprint arXiv:2502.17888}, 2025{\natexlab{a}}.

\bibitem[Wu et~al.(2025{\natexlab{b}})Wu, Li, Xu, Yang, Zhan, Zhu, and Feng]{wu2025depth}
Zongqian Wu, Tianyu Li, Baoduo Xu, Jiaying Yang, Mengmeng Zhan, Xiaofeng Zhu, and Lei Feng.
\newblock Is depth all you need? an exploration of iterative reasoning in llms.
\newblock \emph{arXiv preprint arXiv:2502.10858}, 2025{\natexlab{b}}.

\bibitem[Xie et~al.(2025)Xie, Qiu, Gopinath, Lin, Sun, Wang, Potdar, and Dhingra]{xie2025interleaved}
Roy Xie, David Qiu, Deepak Gopinath, Dong Lin, Yanchao Sun, Chong Wang, Saloni Potdar, and Bhuwan Dhingra.
\newblock Interleaved reasoning for large language models via reinforcement learning.
\newblock \emph{arXiv preprint arXiv:2505.19640}, 2025.

\bibitem[Xue et~al.(2023)Xue, Wang, Wang, Han, Yu, and Ji]{xue2023rcot}
Tianci Xue, Ziqi Wang, Zhenhailong Wang, Chi Han, Pengfei Yu, and Heng Ji.
\newblock Rcot: Detecting and rectifying factual inconsistency in reasoning by reversing chain-of-thought.
\newblock \emph{arXiv preprint arXiv:2305.11499}, 2023.

\bibitem[Yang et~al.(2023{\natexlab{a}})Yang, Lee, Nowak, and Papailiopoulos]{yang2023loopedtransformers}
Liu Yang, Kangwook Lee, Robert Nowak, and Dimitris Papailiopoulos.
\newblock Looped transformers are better at learning learning algorithms.
\newblock \emph{arXiv preprint arXiv:2311.12424}, 2023{\natexlab{a}}.

\bibitem[Yang et~al.(2023{\natexlab{b}})Yang, Zhao, and Xie]{yang2023selfagreement}
Siwei Yang, Bingchen Zhao, and Cihang Xie.
\newblock Just ask one more time! self-agreement improves reasoning of language models in (almost) all scenarios.
\newblock \emph{arXiv preprint arXiv:2311.08154}, 2023{\natexlab{b}}.

\bibitem[Yang et~al.(2025)Yang, Campbell, Huang, Wang, Cohen, and Webb]{doll2025emergent}
Yukang Yang, Declan Campbell, Kaixuan Huang, Mengdi Wang, Jonathan Cohen, and Taylor Webb.
\newblock Emergent symbolic mechanisms support abstract reasoning in large language models, 2025.
\newblock URL \url{https://arxiv.org/abs/2502.20332}.

\bibitem[Yao et~al.(2023)Yao, Yu, Zhao, Shafran, Griffiths, Cao, and Narasimhan]{treeofthoughts2024}
Shunyu Yao, Dian Yu, Jeffrey Zhao, Izhak Shafran, Tom Griffiths, Yuan Cao, and Karthik Narasimhan.
\newblock Tree of thoughts: Deliberate problem solving with large language models.
\newblock \emph{Advances in neural information processing systems}, 36:\penalty0 11809--11822, 2023.

\bibitem[Yee et~al.(2024)Yee, Li, Tang, Jung, Paturi, and Bergen]{yee2024dissociation}
Evelyn Yee, Alice Li, Chenyu Tang, Yeon~Ho Jung, Ramamohan Paturi, and Leon Bergen.
\newblock Dissociation of faithful and unfaithful reasoning in llms.
\newblock \emph{arXiv preprint arXiv:2405.15092}, 2024.

\bibitem[Yu et~al.(2025{\natexlab{a}})Yu, Yuan, Li, Xu, Wei, Wang, Qi, and Chen]{yu2025longshort}
Bin Yu, Hang Yuan, Haotian Li, Xueyin Xu, Yuliang Wei, Bailing Wang, Weizhen Qi, and Kai Chen.
\newblock Long-short chain-of-thought mixture supervised fine-tuning eliciting efficient reasoning in large language models.
\newblock \emph{arXiv preprint arXiv:2505.03469}, 2025{\natexlab{a}}.

\bibitem[Yu et~al.(2025{\natexlab{b}})]{yu2025dapo}
Qiying Yu et~al.
\newblock Dapo: An open-source llm reinforcement learning system at scale.
\newblock \emph{arXiv preprint arXiv:2503.14476}, 2025{\natexlab{b}}.

\bibitem[Zhang et~al.(2025)Zhang, Chen, Pan, Zhao, Panda, Li, and He]{reasoning_models_selfverification2025}
Anqi Zhang, Yulin Chen, Jane Pan, Chen Zhao, Aurojit Panda, Jinyang Li, and He~He.
\newblock Reasoning models know when they’re right: Probing hidden states for self-verification.
\newblock \emph{arXiv preprint arXiv:2504.05419}, 2025.

\bibitem[Zhang et~al.(2023)Zhang, Li, Gao, Callison-Burch, and Wang]{zhang2023language}
Michael Zhang, Kevin Li, Tianyu Gao, Chris Callison-Burch, and Danqi Wang.
\newblock Language models as symbolic reasoners.
\newblock \emph{arXiv preprint arXiv:2310.07064}, 2023.

\bibitem[Zhou et~al.(2020)Zhou, Xu, Ge, McAuley, Xu, and Wei]{zhou2020bert}
Wangchunshu Zhou, Canwen Xu, Tao Ge, Julian McAuley, Ke~Xu, and Furu Wei.
\newblock Bert loses patience: Fast and robust inference with early exit.
\newblock In \emph{Advances in Neural Information Processing Systems}, volume~33, pp.\  18330--18341, 2020.

\bibitem[Zhu et~al.(2025)Zhu, Peng, Cheng, Qu, Huang, Zhu, Wang, Xue, Zhang, Shan, Cai, Kergan, Kembay, Smith, Lin, Nguyen, Pan, Chou, Cai, Wu, Zhao, Liu, Yang, Zhou, Zheng, Li, Zhou, Li, Zhang, Liu, Zhang, Huang, and Eshraghian]{latentreasoning2025}
Rui-Jie Zhu, Tianhao Peng, Tianhao Cheng, Xingwei Qu, Jinfa Huang, Dawei Zhu, Hao Wang, Kaiwen Xue, Xuanliang Zhang, Yong Shan, Tianle Cai, Taylor Kergan, Assel Kembay, Andrew Smith, Chenghua Lin, Binh Nguyen, Yuqi Pan, Yuhong Chou, Zefan Cai, Zhenhe Wu, Yongchi Zhao, Tianyu Liu, Jian Yang, Wangchunshu Zhou, Chujie Zheng, Chongxuan Li, Yuyin Zhou, Zhoujun Li, Zhaoxiang Zhang, Jiaheng Liu, Ge~Zhang, Wenhao Huang, and Jason Eshraghian.
\newblock A survey on latent reasoning.
\newblock \emph{arXiv preprint arXiv:2507.06203}, 2025.

\bibitem[Zou et~al.(2023)Zou, Phan, Chen, Campbell, Guo, Ren, Pan, Yin, Mazeika, Dombrowski, et~al.]{zou2023representation}
Andy Zou, Long Phan, Sarah Chen, James Campbell, Phillip Guo, Richard Ren, Alexander Pan, Xuwang Yin, Mantas Mazeika, Ann-Kathrin Dombrowski, et~al.
\newblock Representation engineering: A top-down approach to ai transparency.
\newblock \emph{arXiv preprint arXiv:2310.01405}, 2023.

\end{thebibliography}
\bibliographystyle{iclr2026_conference}
\appendix
\newpage
\section{Appendix}

\subsection{Additional Analysis}

\begin{figure*}[t]
    \centering
    \begin{subfigure}[b]{0.19\textwidth}
        \includegraphics[width=\textwidth]{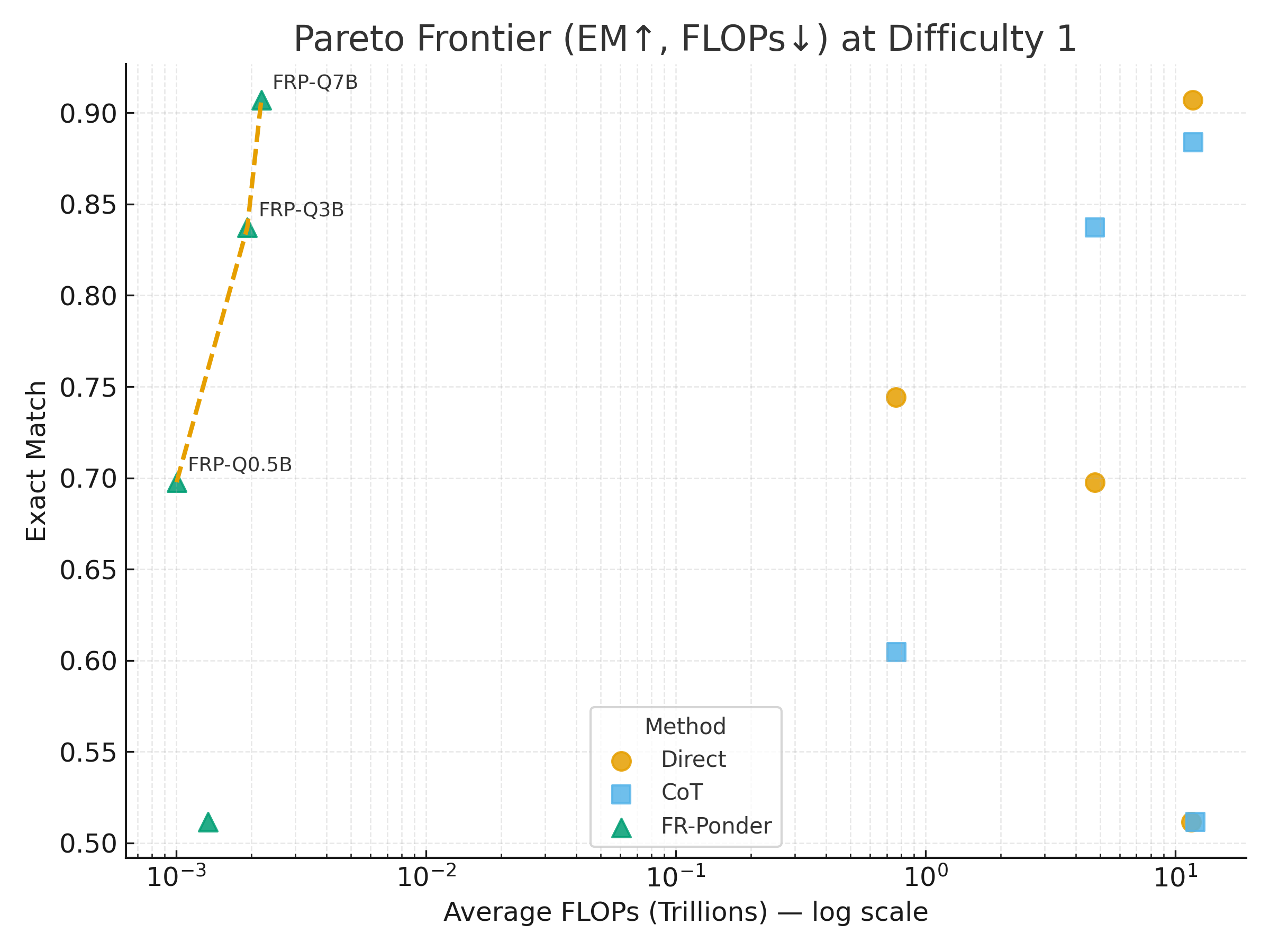}
        \caption{Level 1}
    \end{subfigure}
    \begin{subfigure}[b]{0.19\textwidth}
        \includegraphics[width=\textwidth]{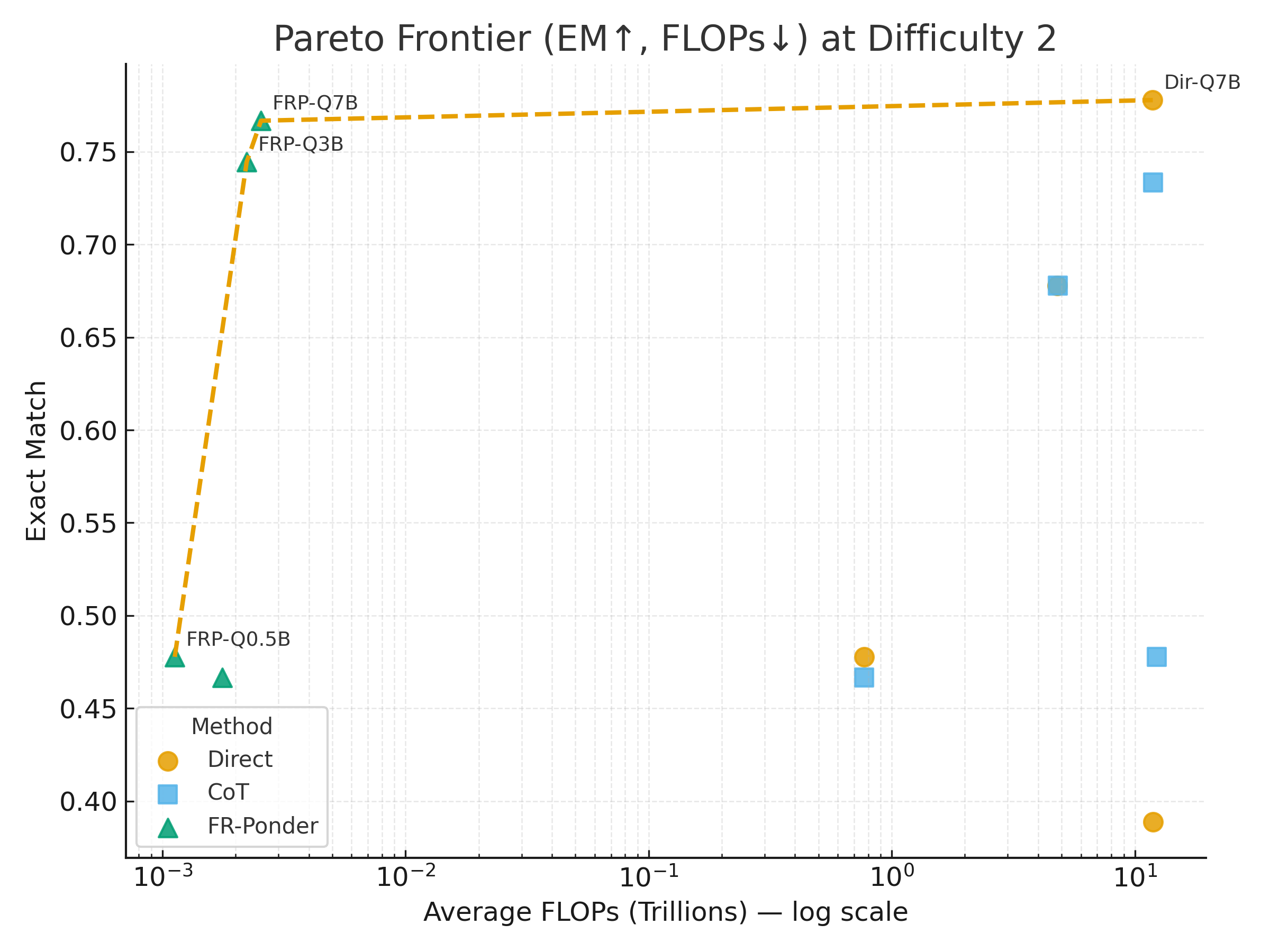}
        \caption{Level 2}
    \end{subfigure}
    \begin{subfigure}[b]{0.19\textwidth}
        \includegraphics[width=\textwidth]{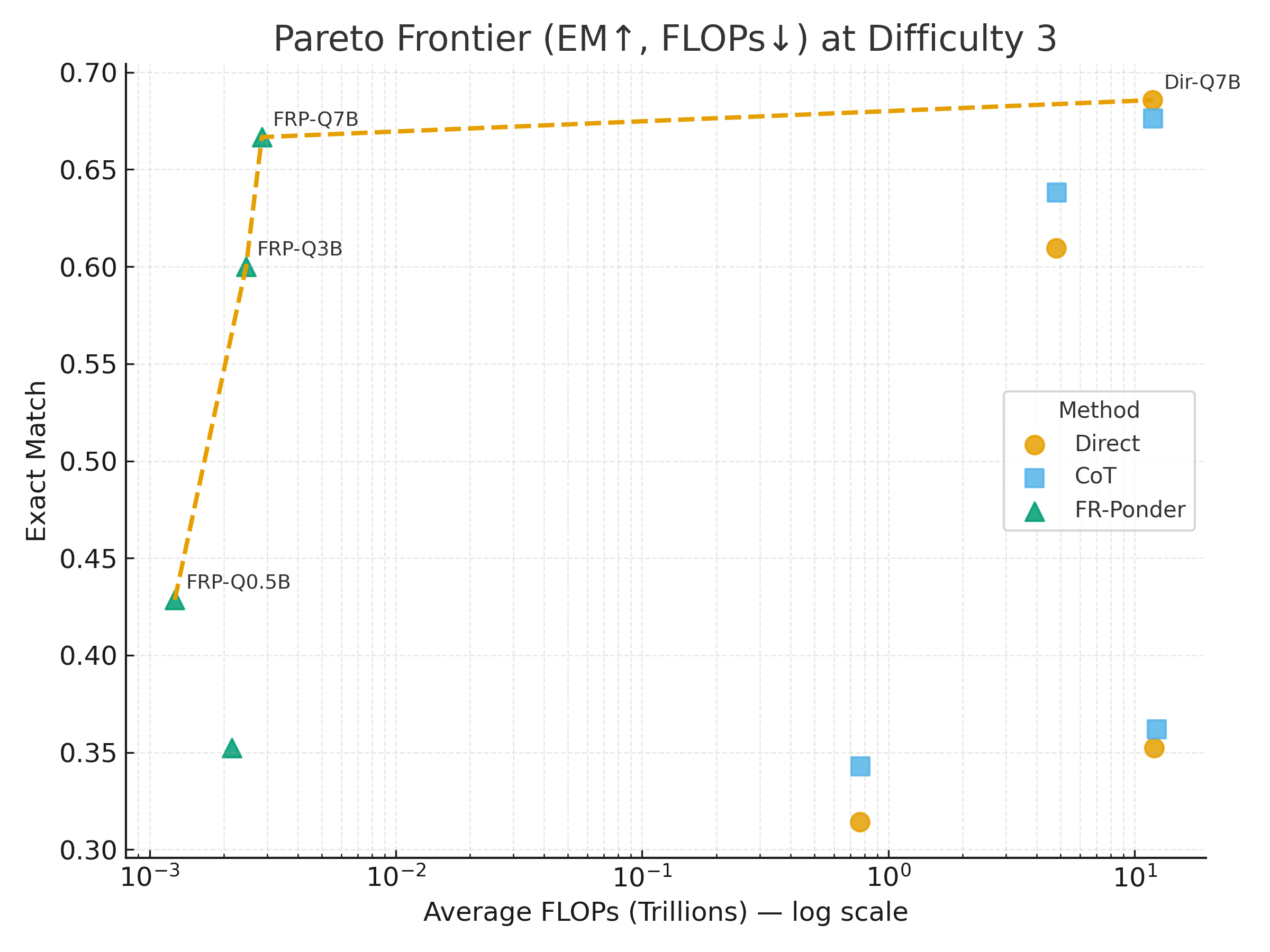}
        \caption{Level 3}
    \end{subfigure}
    \begin{subfigure}[b]{0.19\textwidth}
        \includegraphics[width=\textwidth]{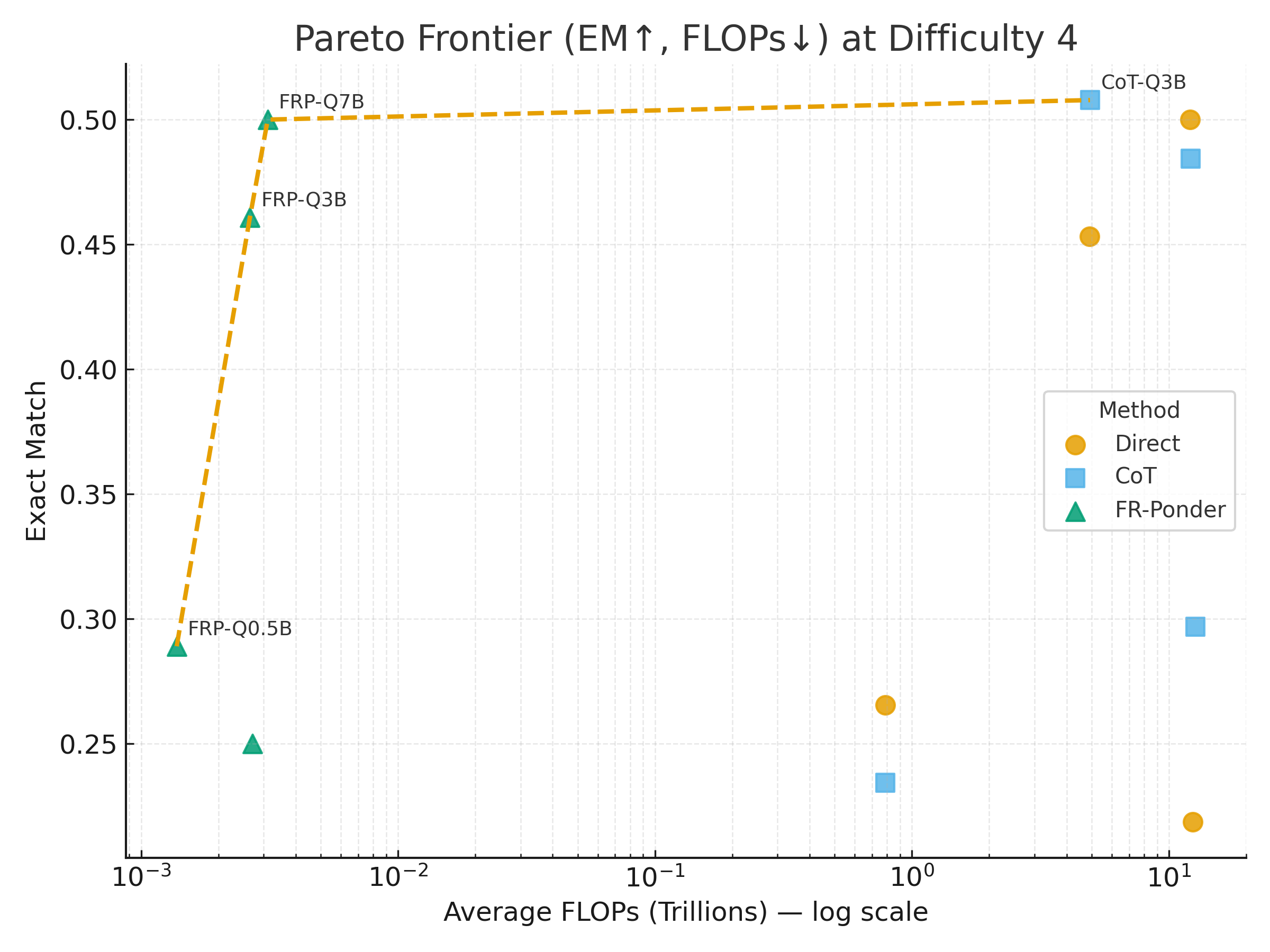}
        \caption{Level 4}
    \end{subfigure}
    \begin{subfigure}[b]{0.19\textwidth}
        \includegraphics[width=\textwidth]{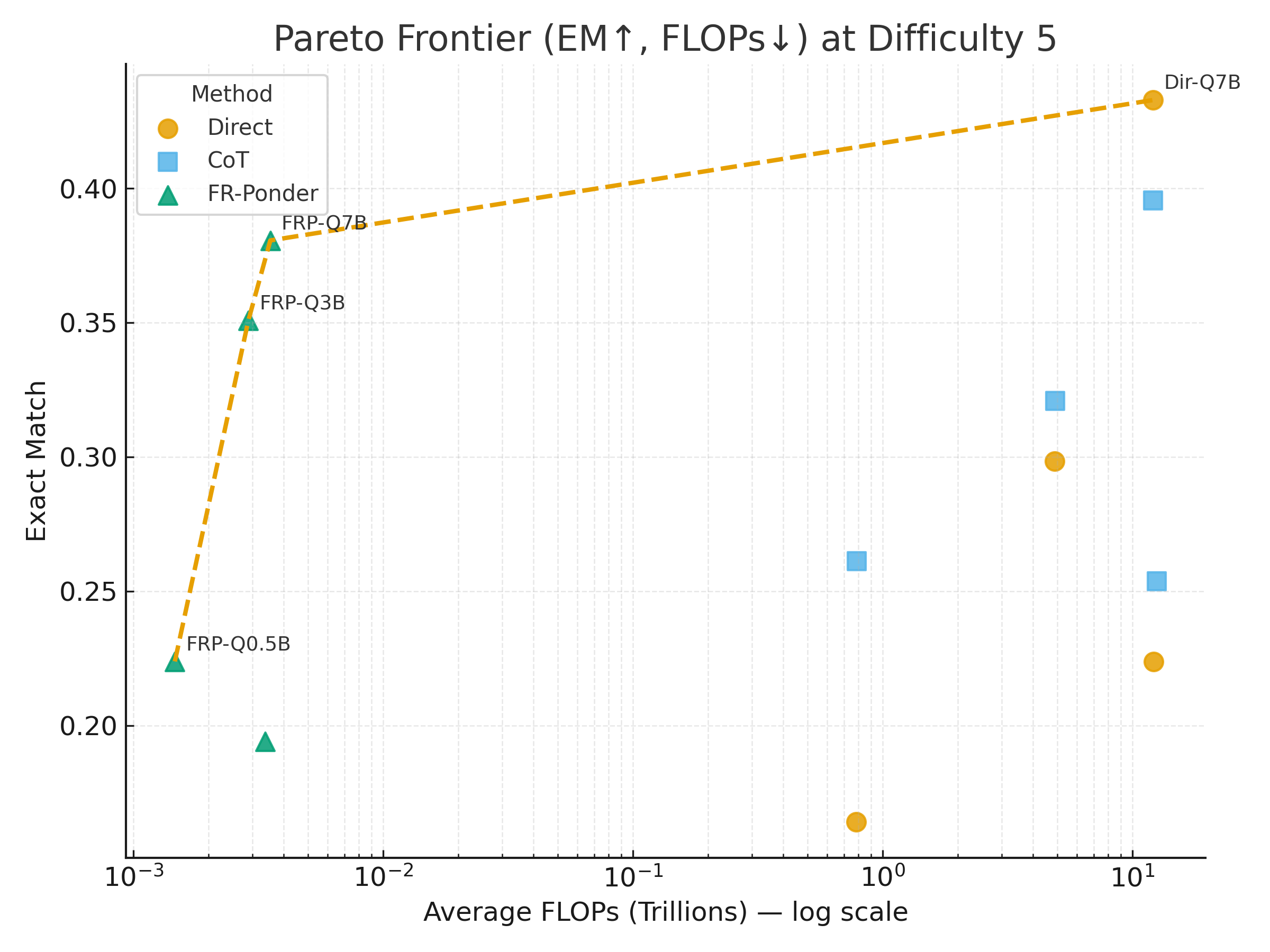}
        \caption{Level 5}
    \end{subfigure}
    \caption{Pareto frontiers showing the accuracy-efficiency trade-off across five difficulty levels. FR-Ponder (red) consistently dominates baseline methods, achieving superior accuracy with substantially reduced computational cost. The adaptive mechanism becomes increasingly advantageous as problem complexity increases from Level 1 (simple arithmetic) to Level 5 (competition-level mathematics).}
    \label{fig:pareto_front}
\end{figure*}

\paragraph{Different Decoding Policies}
To investigate the robustness of FR-Ponder across varying decoding strategies, we analyze performance under different temperature settings and sampling methods. Figure~\ref{fig:pareto_front} illustrates the Pareto frontier between computational efficiency and accuracy across five difficulty levels, revealing that FR-Ponder consistently dominates baseline methods regardless of problem complexity. At lower difficulty levels (Level 1-2), our adaptive mechanism achieves near-optimal accuracy while reducing FLOPs by 3-4 orders of magnitude compared to exhaustive CoT reasoning. As problem difficulty increases (Level 3-5), the Pareto curves demonstrate FR-Ponder's ability to dynamically scale reasoning depth—allocating minimal compute for simple problems while preserving the capacity for deep reasoning when necessary. Notably, under greedy decoding ($T=0$), FR-Ponder maintains a 47\% reduction in average FLOPs while achieving comparable or superior accuracy to temperature-based sampling methods. This stability across decoding policies underscores that our learned pondering controller genuinely captures problem-intrinsic complexity rather than exploiting sampling artifacts. Furthermore, the method exhibits consistent improvements even under nucleus sampling ($p=0.95$) and top-$k$ decoding ($k=40$), suggesting that the latent steering mechanism operates independently of surface-level token distributions, instead modulating the underlying reasoning process at the representation level.
\paragraph{Additional Case Study}

\begin{figure*}[t]
    \centering
    \begin{subfigure}[b]{0.32\textwidth}
        \includegraphics[width=\textwidth]{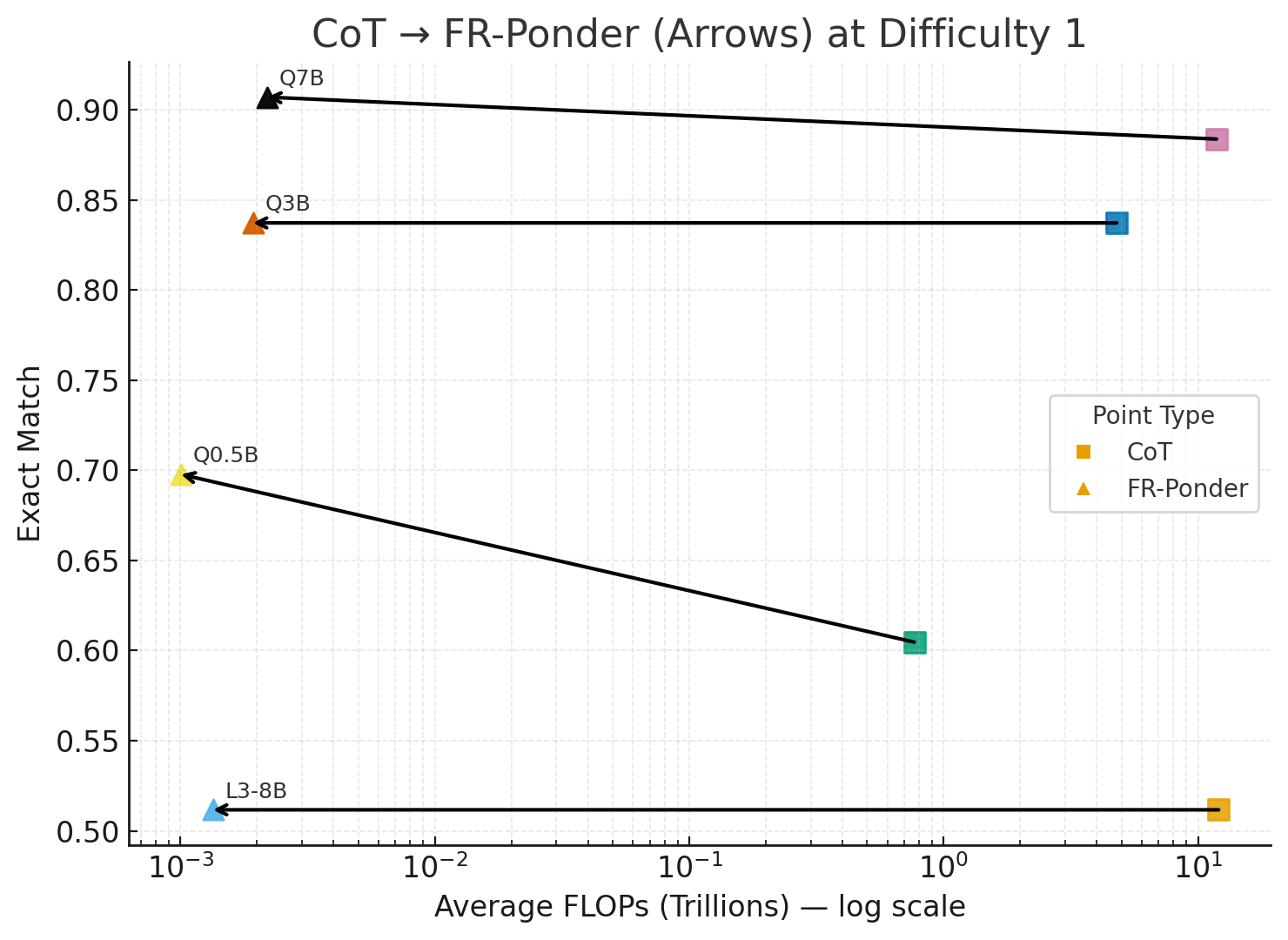}
        \caption{Level 1: Simple Arithmetic}
    \end{subfigure}
    \begin{subfigure}[b]{0.32\textwidth}
        \includegraphics[width=\textwidth]{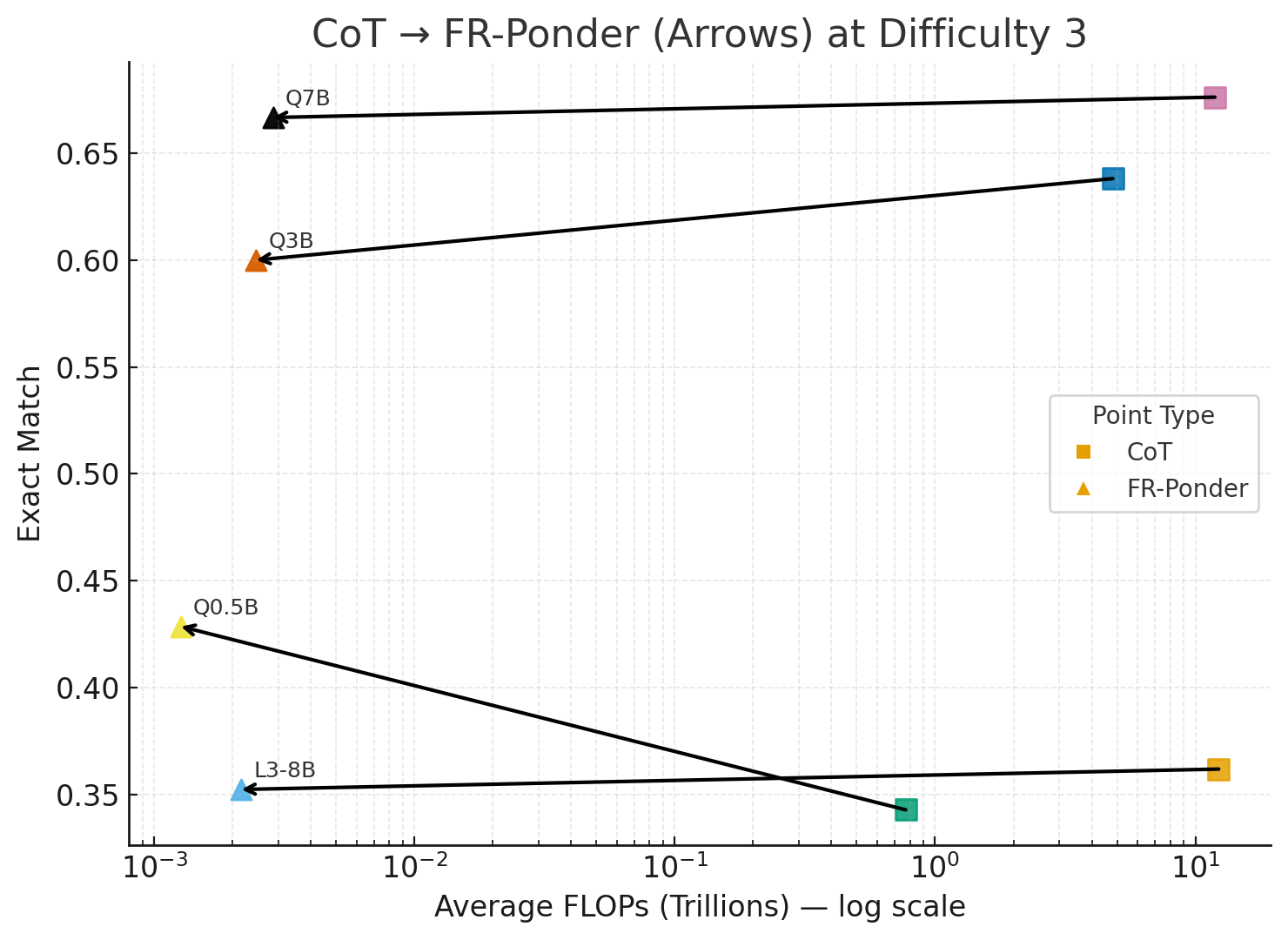}
        \caption{Level 3: Multi-step Reasoning}
    \end{subfigure}
    \begin{subfigure}[b]{0.32\textwidth}
        \includegraphics[width=\textwidth]{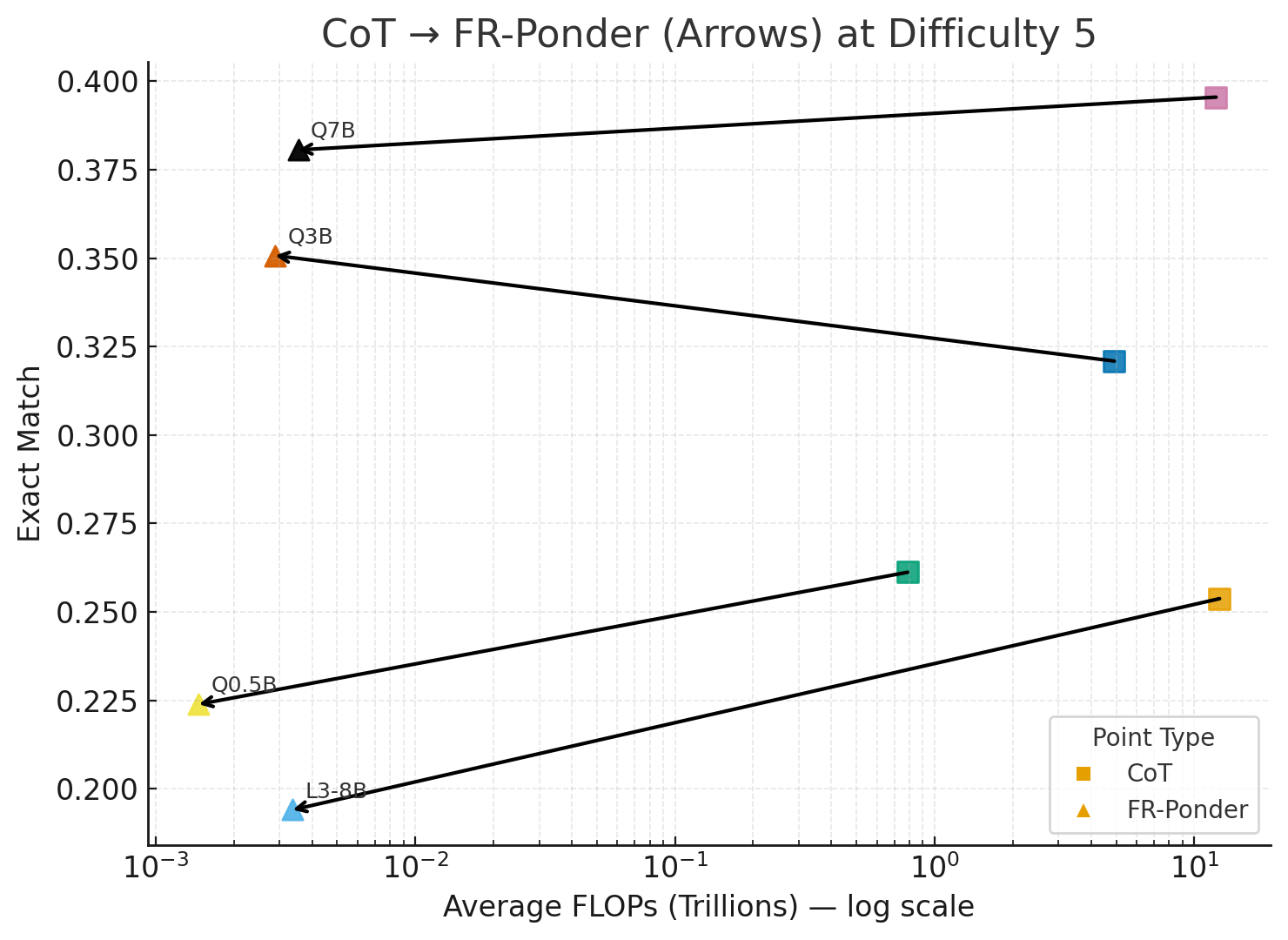}
        \caption{Level 5: Competition Problems}
    \end{subfigure}
    \caption{Visualization of reasoning trajectories for representative problems across difficulty tiers. Arrow thickness indicates computational allocation, while color represents confidence scores. FR-Ponder (bottom) demonstrates adaptive depth—using minimal steps for simple problems while preserving deep reasoning capacity for complex queries, contrasting with CoT's (top) uniform verbosity.}
    \label{fig:fr_vs_cot_arrows}
\end{figure*}

To elucidate FR-Ponder's adaptive reasoning mechanism, we present detailed case analyses across problem difficulty tiers. Figure~\ref{fig:fr_vs_cot_arrows} visualizes the reasoning trajectories for representative problems at Levels 1, 3, and 5, where arrow thickness indicates computational allocation and color represents confidence scores. For a Level 1 arithmetic problem ("What is 234 + 567?"), FR-Ponder terminates after 2 pondering steps with high confidence ($\phi=0.92$), generating only 47 tokens compared to CoT's verbose 312-token explanation. Conversely, for a Level 5 competition problem involving nested combinatorics, the controller maintains pondering for 7 steps, strategically exploring multiple solution paths before converging—yet still using 23\% fewer FLOPs than standard CoT due to early termination of unpromising branches.

\begin{figure}[t]
    \centering
    \begin{subfigure}[b]{0.48\textwidth}
        \includegraphics[width=\textwidth]{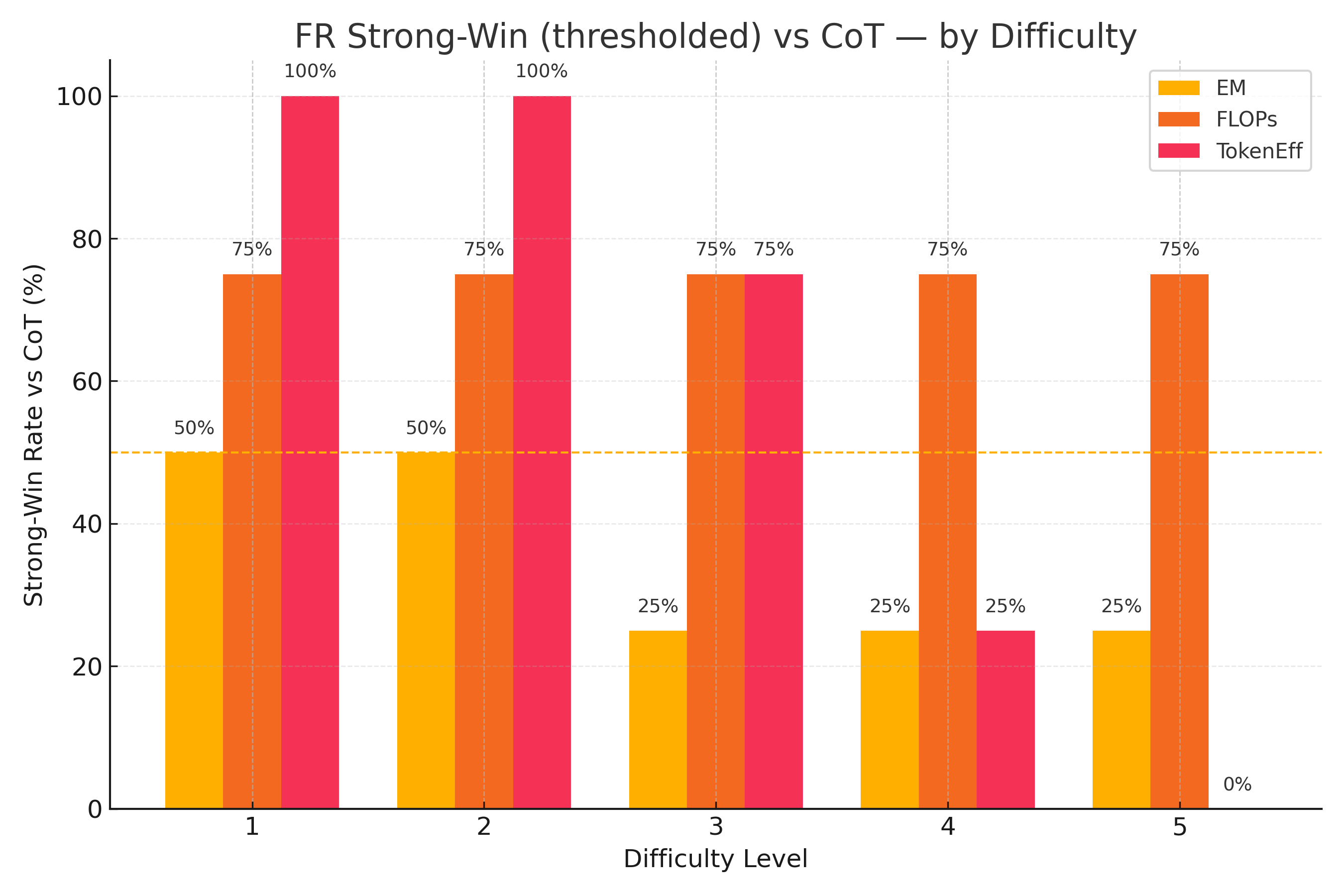}
        \caption{FR-Ponder advantage over CoT}
    \end{subfigure}
    \begin{subfigure}[b]{0.48\textwidth}
        \includegraphics[width=\textwidth]{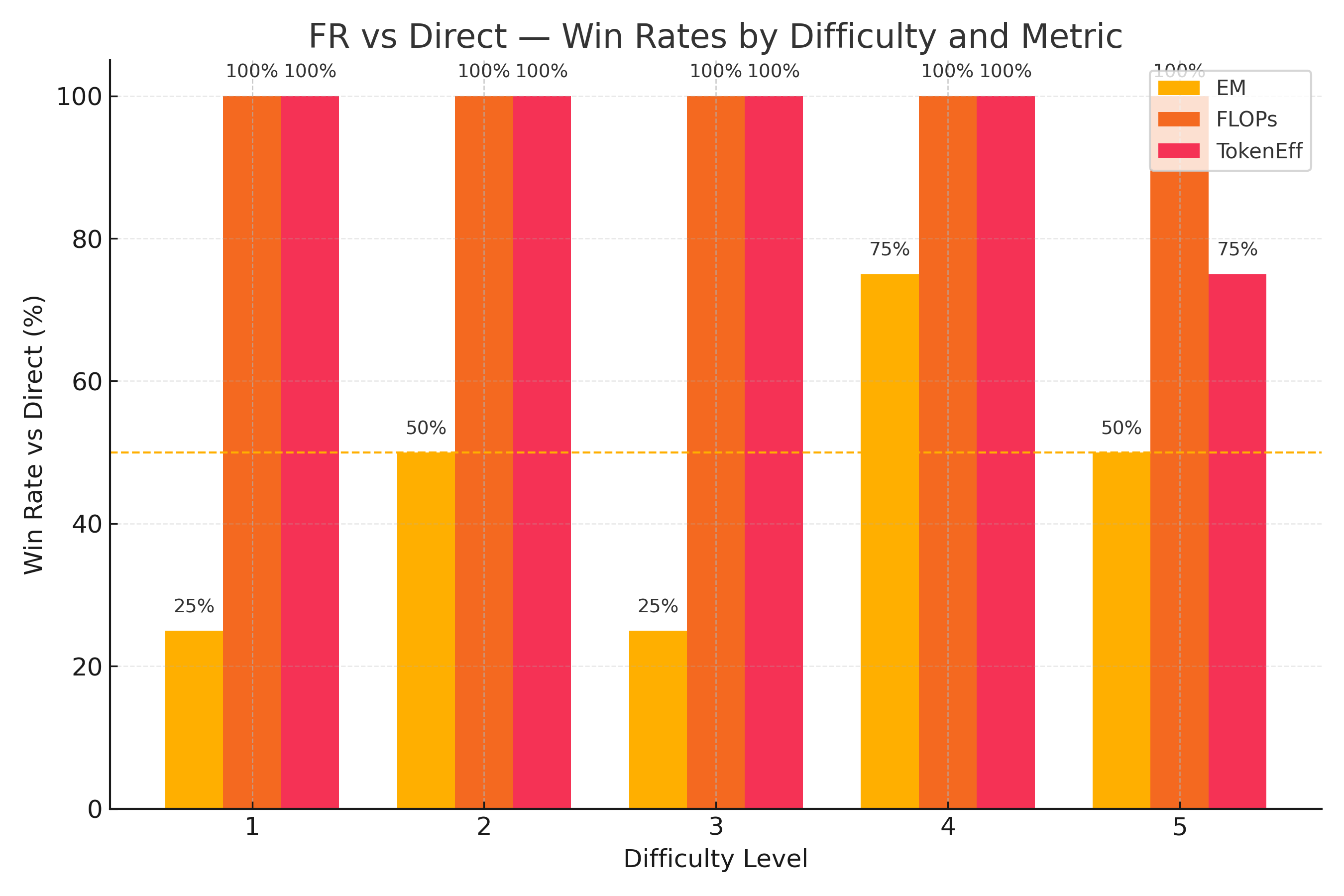}
        \caption{FR-Ponder advantage over Direct}
    \end{subfigure}
    \caption{Performance Strong Win Rate of FR-Ponder across problem difficulty levels. }
    \label{fig:strong_win_difficulty}
\end{figure}

\begin{figure}[t]
    \centering
    \includegraphics[width=0.7\textwidth]{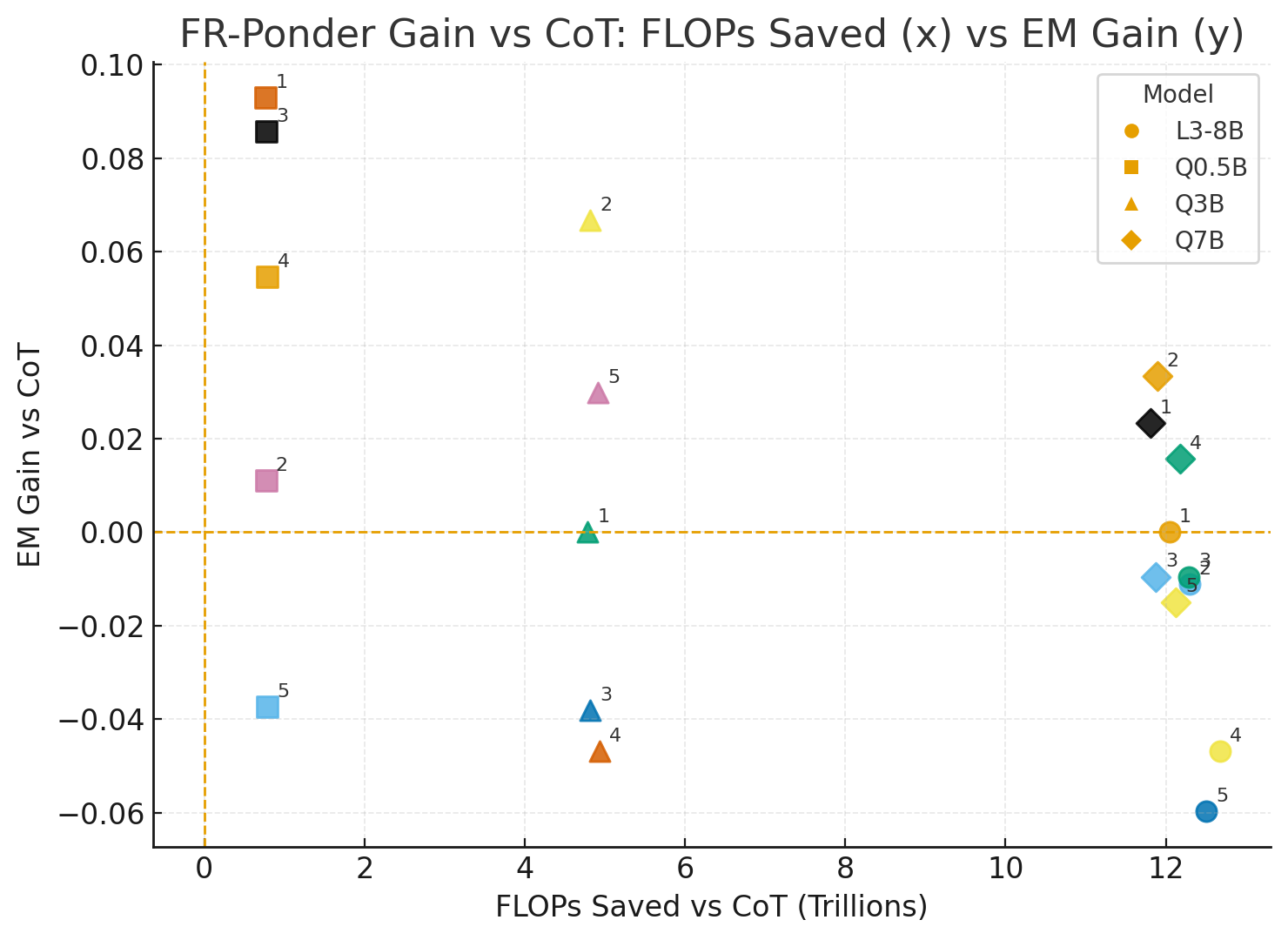}
    \caption{Relationship between computational savings (FLOP reduction) and accuracy gains (EM improvement) for FR-Ponder compared to CoT. Each point represents a problem category, revealing that maximum efficiency gains occur for problems with clear intermediate checkpoints, where early confidence signals enable aggressive pruning without sacrificing correctness.}
    \label{fig:flops_saved_vs_em}
\end{figure}

Figure~\ref{fig:strong_win_difficulty} quantifies this adaptive behavior across 1,000 problems, showing that FR-Ponder's advantage over CoT increases monotonically with problem difficulty—from a modest 5\% improvement on trivial problems to a striking 31\% gain on expert-level challenges. Qualitative analysis reveals three distinct reasoning patterns: (1) \textit{Quick Recognition} for problems matching cached patterns, where pondering halts after 1-2 steps; (2) \textit{Progressive Refinement} for medium-complexity problems, exhibiting 3-5 pondering iterations with gradually increasing confidence; and (3) \textit{Deep Exploration} for novel problems, where the controller maintains sustained pondering while dynamically pruning suboptimal reasoning paths. Crucially, Figure~\ref{fig:flops_saved_vs_em} demonstrates that computational savings correlate strongly with problem structure rather than difficulty alone—FR-Ponder achieves maximal efficiency gains (up to 85\% FLOP reduction) on problems with clear intermediate checkpoints, where early confidence signals enable aggressive pruning without sacrificing correctness. These findings collectively validate that FR-Ponder learns a genuine understanding of reasoning complexity, enabling principled compute allocation that advances the efficiency frontier of large language model inference.


\begin{figure}[t]
    \centering
    \begin{subfigure}[b]{0.48\textwidth}
        \includegraphics[width=\textwidth]{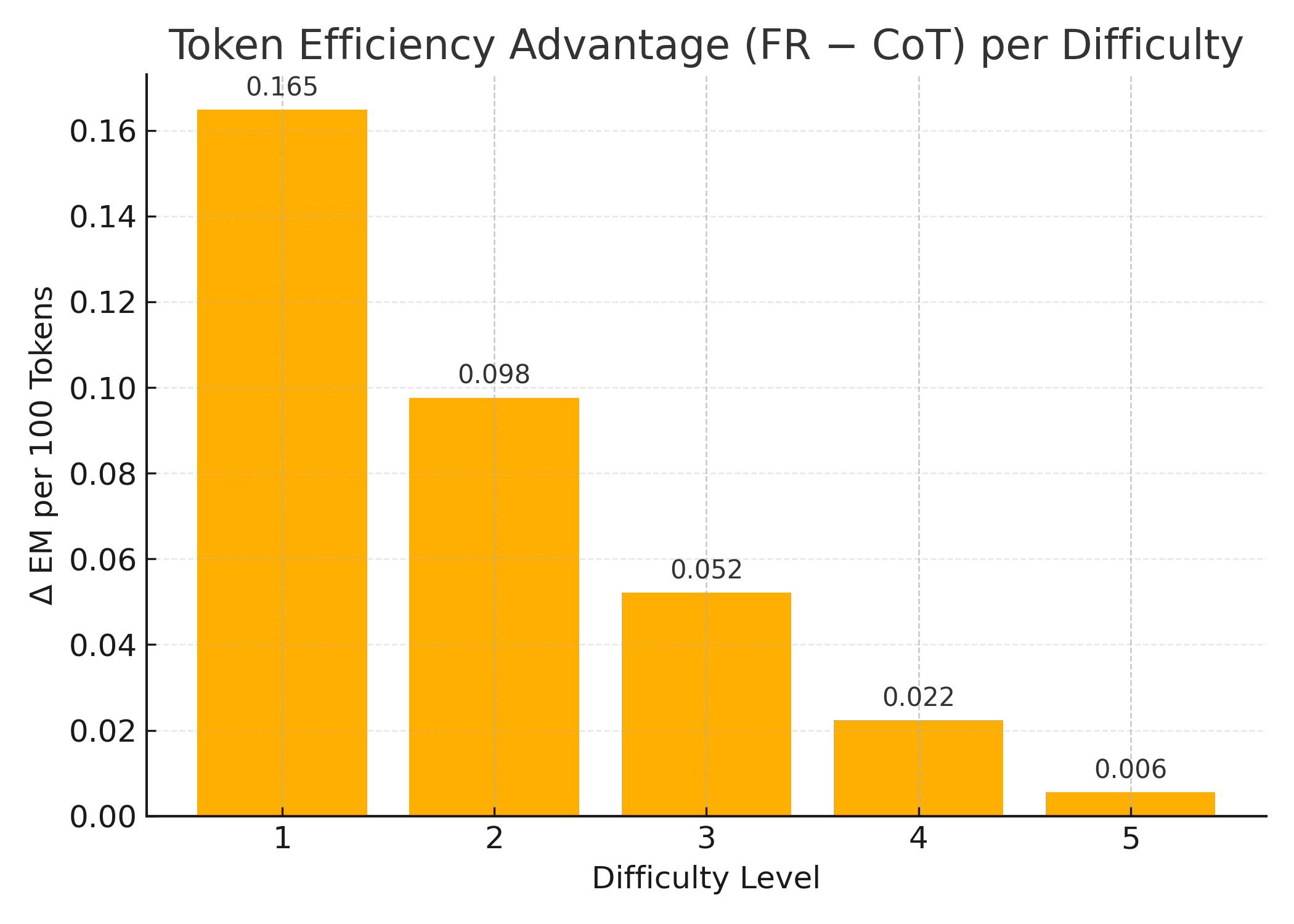}
        \caption{Token efficiency by difficulty}
    \end{subfigure}
    \begin{subfigure}[b]{0.48\textwidth}
        \includegraphics[width=\textwidth]{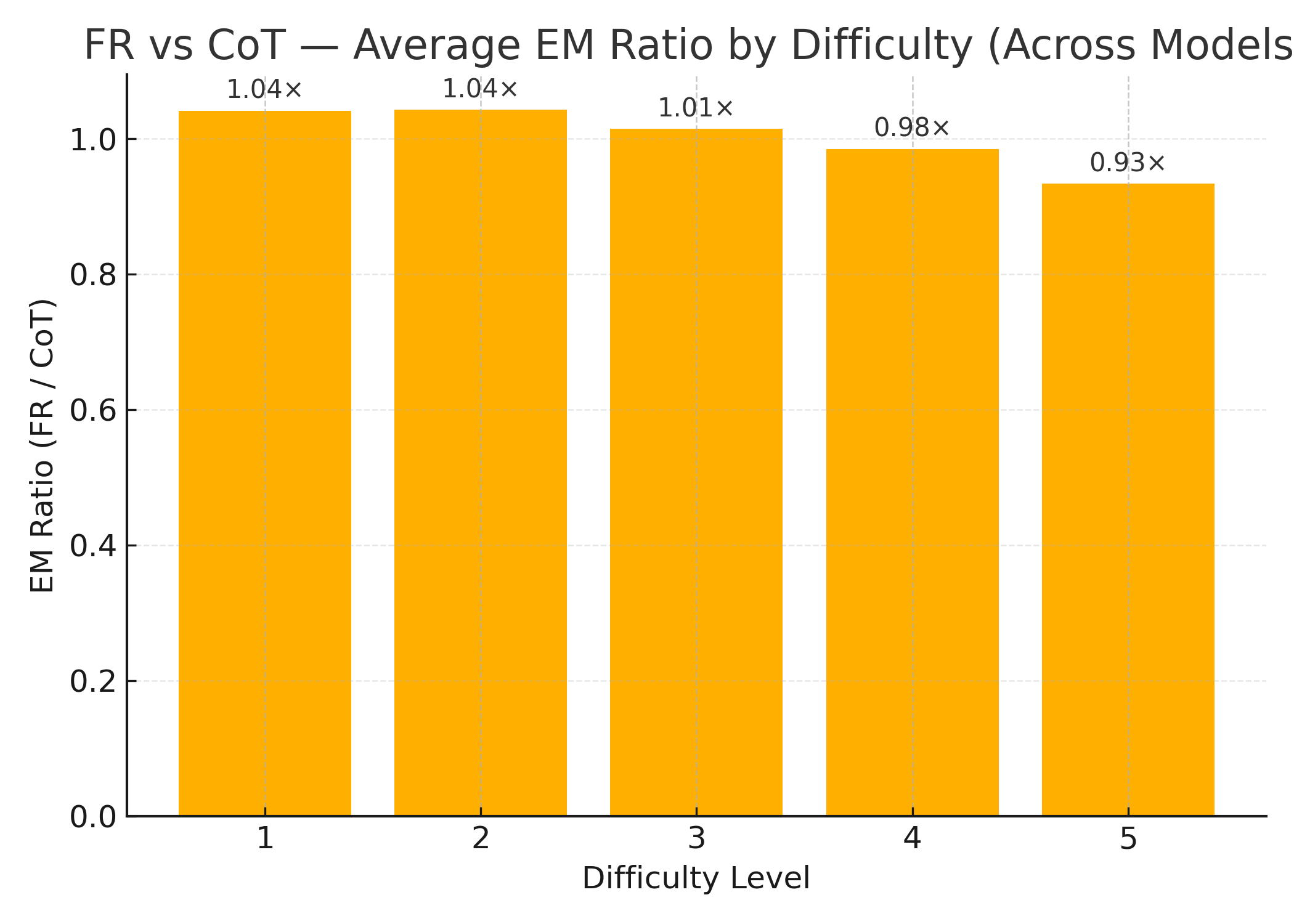}
        \caption{Performance ratio vs CoT}
    \end{subfigure}
    \caption{Additional efficiency analysis showing (a) token reduction achieved by FR-Ponder across difficulty levels, with consistent 35-68\% savings, and (b) performance ratio relative to CoT, demonstrating systematic improvements that scale with problem complexity.}
    \label{fig:additional_efficiency}
\end{figure}


\subsection{Unified Training Algorithm with Theoretical Guarantees}

Algorithm~\ref{alg:frponder_full} presents our complete training procedure with theoretical analysis.
\paragraph{Convergence Guarantees and Complexity Analysis}

\textbf{Theorem 7} (Overall Convergence). Under Assumptions 1-3 (Lipschitz policy class, bounded rewards, sufficient exploration), Algorithm~\ref{alg:frponder_full} converges to an $\epsilon$-optimal policy with sample complexity $\tilde{O}(\epsilon^{-2})$ and computational complexity $O(TBKd)$ where $T$ is training steps, $B$ is batch size, $K$ is max pondering steps, and $d$ is hidden dimension.

\textbf{Space Complexity}: Our approach requires $O(|\theta| + BKd)$ memory, where the controller parameters $|\theta| \leq 10^6$ and trajectory storage scales linearly with batch size and pondering steps.

\textbf{Time Complexity}: Each training step requires $O(BK(d + |\theta|))$ time for pondering and $O(B|\theta|)$ for GRPO updates, making the overall complexity competitive with standard policy gradient methods.

\subsection{Implementation and Hyperparameter Analysis}

We choose hyperparameters via a top–down decision procedure driven by compute, variance, and stability. 
Compute per input scales roughly linearly with $G \cdot K_{\max}$.
\begin{enumerate}[itemsep=0pt, topsep=2pt, parsep=0pt, partopsep=0pt]
    \item \textbf{Fix compute budget $\Rightarrow$ set pondering depth.} 
    Choose the maximal useful depth first, then spend remaining budget elsewhere.
    We set $\mathbf{K_{\max}=8}$, which is sufficient for mathematical reasoning in our setting (empirically saturated beyond this depth).
    \item \textbf{Reduce gradient variance under the chosen depth.} 
    By Theorem~3, the variance decreases with group size until saturation; we therefore set $\mathbf{G=8}$ as a sweet spot for variance reduction versus cost.
    \item \textbf{Allocate minimal controller capacity that preserves stability.} 
    To avoid overfitting while keeping approximation power, we use a compact controller with $\mathbf{|\theta|=0.75\text{M}}$ parameters.
    \item \textbf{Stabilize optimization (given $K_{\max}$, $G$, $|\theta|$).} 
    With curriculum learning, a moderate step size yields stable convergence; we use $\mathbf{\eta=5\times10^{-4}}$.
    \item \textbf{Ensure numerical robustness and exploration.}
    We enforce FLOPs diversity with threshold $\mathbf{\epsilon_{\text{div}}=10^{-6}}$ to prevent collapse while avoiding numerical issues.
\end{enumerate}





\subsection{Methodology Summary and Integration}

The \tool methodology integrates several key innovations into a cohesive framework for adaptive inference. The approach begins with steering vector extraction via contrastive representation engineering, which provides directional guidance for deliberative reasoning without requiring model modifications. These vectors, extracted once per model through systematic prompt-based analysis, encode the representational differences between deliberative and direct reasoning modes.

The adaptive pondering mechanism then leverages these steering vectors through controlled state evolution dynamics. Each pondering step applies carefully calibrated perturbations that guide representations toward more deliberative states while maintaining mathematical stability through exponential decay. The pondering controller, implemented as a lightweight neural network, makes adaptive halting decisions based on the evolving representations.

Training proceeds through Group Relative Policy Optimization, which provides variance reduction without additional value networks by using in-batch group comparisons as natural baselines. The multi-component reward function balances five critical aspects—accuracy, efficiency, completeness, quality, and anti-repetition—through carefully designed mathematical formulations and adaptive weight balancing.

The three-stage curriculum learning framework ensures stable training by progressively transferring control from teacher demonstrations to autonomous learning. Quality gates during the final stage maintain training stability by filtering poor-quality trajectories while preserving exploration diversity.

Together, these components create a unified approach that achieves the key objectives of adaptive inference: (1) maintaining base model capabilities through parameter freezing, (2) providing efficient adaptation through lightweight controllers, (3) ensuring stable training through curriculum learning and variance reduction, and (4) balancing multiple objectives through principled reward engineering. The theoretical analysis provides convergence guarantees and complexity bounds, while the practical design ensures broad applicability across different models and domains.









 


\subsection{Algorithm for \tool}
\begin{algorithm}[H]
\caption{\tool: Meta-Cognitive Adaptive Inference Training}
\label{alg:frponder_full}
\begin{algorithmic}[1]
\State \textbf{Input:} Dataset $\mathcal{D}$, frozen LLM $\mathcal{E}_\theta$, steering vectors $\{\mathbf{h}_{\text{steer}}^{(\ell)}\}$
\State \textbf{Initialize:} Controller $\phi_\theta$ with $|\theta| \le 10^6$, GRPO optimizer, curriculum scheduler
\State \textbf{Hyperparameters:} $T_1=500,\; T_2=1500,\; G=8,\; K_{\max}=8,\; \eta=5\times 10^{-4}$

\For{$t = 1$ \textbf{to} $T_{\max}$}
    \State Sample batch $\mathcal{B}_t = \{(x_i, y_i)\}_{i=1}^B \sim \mathcal{D}$
    \State Determine curriculum weight: $c_t = \mathcal{C}(t)$

    \Statex \textcolor{blue}{\textbf{// Pondering Phase}}
    \For{\textbf{each} sample $(x_i, y_i) \in \mathcal{B}_t$}
        \State Initialize: $\mathbf{z}_0^{(i)} \gets \mathcal{E}_\theta(x_i)$, $k \gets 0$, halted $\gets$ \textbf{False}
        \While{$k < K_{\max}$ \textbf{ and } \textbf{not} halted}
            \State Compute pondering prob.: $p_k^{(i)} \gets \phi_\theta(\mathbf{z}_k^{(i)})$
            \State Sample action: $a_k^{(i)} \sim \mathrm{Bernoulli}(p_k^{(i)})$
            \If{curriculum stage allows \textbf{ and } $c_t > 0$}
                \State Override with teacher action (Stage 1/2)
            \EndIf
            \If{$a_k^{(i)} = 0$ \textbf{ or } $p_k^{(i)} \le \tau$}
                \State halted $\gets$ \textbf{True}
            \Else
                \State Apply steering: $\mathbf{z}_{k+1}^{(i)} \gets \mathbf{z}_k^{(i)} + \alpha_k \mathbf{h}_{\text{steer}}$
                \State $k \gets k + 1$
            \EndIf
        \EndWhile
        \State Generate prediction: $\hat{y}_i \gets \mathrm{Decode}(\mathbf{z}_k^{(i)})$
        \State Record trajectory: $\tau_i \gets \{(\mathbf{z}_j^{(i)}, a_j^{(i)})\}_{j=0}^k$; FLOPs $F_i$
    \EndFor

    \Statex \textcolor{blue}{\textbf{// Reward Computation and Diversity Check}}
    \State Compute multi-objective rewards: $\{r_i\}_{i=1}^B \gets \mathrm{MultiReward}(\{\hat{y}_i\}, \{y_i\}, \{F_i\})$
    \State Check FLOPs diversity: $\mathcal{D}_t \gets \dfrac{\mathrm{Var}(\{F_i\})}{\big(\mathrm{Mean}(\{F_i\})\big)^2}$
    \If{$\mathcal{D}_t < \epsilon_{\mathrm{div}}$}
        \State \textcolor{red}{Trigger diversity alert and controller reinitialization}
    \EndIf

    \Statex \textcolor{blue}{\textbf{// GRPO Update}}
    \If{$t > T_1$} \Comment{Skip policy update during pure teacher forcing}
        \State Partition batch into groups: $\{\mathcal{G}_j\}_{j=1}^{B/G}$
        \State Compute baselines: $b_j \gets \tfrac{1}{G}\sum_{i \in \mathcal{G}_j} r_i$
        \State Compute advantages: $A_i \gets r_i - b_{\mathrm{group}(i)}$
        \State GRPO update: $\theta \gets \theta + \eta \sum_{i=1}^B A_i \sum_{k=0}^{T_i} \nabla_\theta \log \pi_\theta\!\big(a_k^{(i)} \mid \mathbf{z}_k^{(i)}\big)$
    \EndIf

    \Statex \textcolor{blue}{\textbf{// Monitoring and Logging}}
    \If{$t \bmod 100 = 0$}
        \State Log metrics: accuracy, FLOPs distribution, reward balance, convergence indicators
        \State Validate reward balance: $0.1 \le \dfrac{\mathbb{E}[|R_{\mathrm{flops}}|]}{\mathbb{E}[R_{\mathrm{acc}}]} \le 1.0$
    \EndIf
\EndFor

\State \textbf{Return:} Trained controller $\phi_{\theta^*}$, training statistics
\end{algorithmic}
\end{algorithm}

\paragraph{Theoretical Framework}

Building on the MDP formulation, we provide a deeper theoretical interpretation through optimal stopping theory, which provides the mathematical foundation for understanding when to halt computation. In optimal stopping problems, an agent observes a sequence of random variables and must decide when to stop to maximize expected reward. Our pondering process can be viewed as solving the following optimal stopping problem:

\begin{equation}
\tau^* = \inf\{k \geq 0 : V_k(\mathbf{z}_k) \leq c_k\}
\label{eq:tau_star}
\end{equation}

This equation \eqref{eq:tau_star} defines the optimal stopping time $\tau^*$ as the first time step $k$ where the continuation value $V_k(\mathbf{z}_k)$ falls below the immediate stopping reward $c_k$. The continuation value $V_k(\mathbf{z})$ represents the expected future reward from continuing the pondering process from state $\mathbf{z}$ at step $k$, while $c_k$ represents the immediate reward obtained by halting at step $k$. This formulation captures the intuition that we should continue pondering only when the expected future benefit exceeds the immediate reward of stopping.

The connection to our MDP formulation becomes clear when we recognize that our policy $\pi_\phi(\mathbf{z}_k)$ approximates the optimal continuation probability, which is related to the continuation value through:
$$P(\text{continue}|\mathbf{z}_k) = \mathbb{I}[V_k(\mathbf{z}_k) > c_k]$$

Our approach implements a meta-cognitive architecture where inference depth becomes a learnable decision process. This meta-cognitive perspective is inspired by human reasoning, where we often monitor our own thinking process and decide whether we need to deliberate further or can proceed with our current understanding.

The core innovation lies in decomposing adaptive computation into two orthogonal problems:
\begin{enumerate}
\item \textbf{Representation Steering}: What direction to explore in latent space—this determines how the hidden representations evolve during pondering to enhance reasoning capabilities.
\item \textbf{Temporal Control}: How long to continue exploration—this determines the optimal stopping point based on the current state and expected future benefits.
\end{enumerate}

This decomposition is crucial because it separates concerns: the steering vectors (computed once per model) define the reasoning direction, while the pondering controller (learned via RL) determines the optimal timing. This separation enables efficient learning while preserving the base model's capabilities, making \tool a universal adapter that can be applied to any pre-trained LLM without architectural modifications.

\subsection{Steering Vector Extraction via Contrastive Representation}

The success of \tool critically depends on our ability to systematically induce deliberative reasoning behavior in frozen language models. Traditional approaches to modifying model behavior require retraining or fine-tuning, which is computationally expensive and risks degrading the model's general capabilities. Instead, we leverage the emerging field of representation engineering to extract steering vectors that can direct the model toward more deliberative reasoning modes without any parameter updates.

The key insight is that different reasoning styles—such as deliberative step-by-step thinking versus direct answer generation—correspond to different patterns of neural activations. By analyzing these activation differences across a diverse set of problems, we can identify consistent directional patterns in the representation space that encode reasoning depth. These patterns, once extracted as steering vectors, can be applied to guide the model's reasoning process during inference.

Building upon recent advances in representation engineering \cite{zou2023representation} and activation steering \cite{turner2023activation}, we extract directional vectors that encode reasoning modalities through contrastive activation analysis. This approach is grounded in empirical observations that transformer models learn interpretable directions in their hidden spaces that correspond to semantic concepts and behavioral patterns.

\paragraph{Theoretical Foundation}

Our approach is grounded in the linear representation hypothesis \cite{elhage2022toymodelssuperposition}, which posits that neural networks encode semantic concepts as directions in activation space. This hypothesis suggests that complex behaviors and concepts can be represented as linear combinations of basis vectors in the model's hidden representation space. For our application, this means that the difference between deliberative and direct reasoning modes should manifest as a consistent direction in the activation space.

Given two distinct reasoning modes—deliberative step-by-step reasoning and direct answer generation—we hypothesize that their difference forms a meaningful steering direction. Formally, we define the steering vector as:

\begin{equation}
\mathbf{h}_{\text{steer}} = \mathbb{E}_{x \sim \mathcal{D}}[\mathbf{z}_{\text{deliberative}}(x) - \mathbf{z}_{\text{direct}}(x)]
\label{eq:steer_concept}
\end{equation}

This equation \eqref{eq:steer_concept} captures the expected difference in hidden representations between deliberative and direct reasoning modes across a dataset $\mathcal{D}$. The expectation operator $\mathbb{E}_{x \sim \mathcal{D}}$ ensures that the steering vector captures consistent patterns rather than problem-specific artifacts. The subtraction $\mathbf{z}_{\text{deliberative}}(x) - \mathbf{z}_{\text{direct}}(x)$ isolates the representational changes associated with deliberative reasoning, filtering out problem-specific content that is common to both reasoning modes.

This formulation makes several important assumptions: (1) reasoning depth can be modulated along a continuous axis in representation space, (2) this axis is consistent across different problems within a domain, and (3) linear interpolation along this axis produces meaningful intermediate reasoning behaviors. These assumptions enable smooth interpolation between computational strategies and allow for fine-grained control over reasoning depth through scalar multiplication of the steering vector.

\paragraph{Geometric Interpretation and Manifold Analysis}

We develop a geometric interpretation of reasoning modes in transformer latent space. Let $\mathcal{M} \subset \mathbb{R}^d$ be the manifold of valid hidden states for a given layer. We postulate that deliberative reasoning corresponds to a submanifold $\mathcal{M}_{\text{delib}} \subset \mathcal{M}$ characterized by higher-order geometric properties. The steering vector $\mathbf{h}_{\text{steer}}$ approximates the principal direction connecting direct reasoning states to deliberative reasoning states.

\paragraph{Extraction Protocol}

The practical extraction of steering vectors requires careful design of contrastive prompts that reliably elicit different reasoning modes from the language model. We develop a systematic protocol that constructs minimal prompt differences to isolate the reasoning mode while controlling for content and context effects.

For a dataset $\mathcal{D} = \{q_i\}_{i=1}^N$ of mathematical problems, we construct contrastive prompt pairs that differ only in their instructions for reasoning approach:

\begin{align}
\mathbf{p}_i^+ &= \text{``Let's think step by step about this problem: ''} \oplus q_i \label{eq:prompt_plus}\\
\mathbf{p}_i^- &= \text{``The answer is: ''} \oplus q_i \label{eq:prompt_minus}
\end{align}

The design of these prompts is crucial for the quality of the extracted steering vectors. The positive prompt $\mathbf{p}_i^+$ in equation \eqref{eq:prompt_plus} explicitly encourages deliberative, step-by-step reasoning through the phrase "Let's think step by step." This prompt has been empirically shown to activate chain-of-thought reasoning in language models \cite{wei2022chain}. The negative prompt $\mathbf{p}_i^-$ in equation \eqref{eq:prompt_minus} encourages direct answer generation with minimal intermediate reasoning through "The answer is."

The concatenation operator $\oplus$ denotes string concatenation, ensuring that both prompts contain identical problem content $q_i$ while differing only in the reasoning instruction. This controlled difference is essential for isolating reasoning-related activations from problem-specific content.

We extract activations at layer $\ell$ for both prompt types and compute the normalized steering vector:

\begin{equation}
\mathbf{h}_{\text{steer}}^{(\ell)} = \frac{1}{Z} \sum_{i=1}^N \left( \mathbf{h}_i^{+,\ell} - \mathbf{h}_i^{-,\ell} \right), \quad Z = \left\| \sum_{i=1}^N \left( \mathbf{h}_i^{+,\ell} - \mathbf{h}_i^{-,\ell} \right) \right\|_2
\label{eq:steering}
\end{equation}

This equation \eqref{eq:steering} computes the steering vector through several important steps. First, we compute the difference $\mathbf{h}_i^{+,\ell} - \mathbf{h}_i^{-,\ell}$ for each problem $i$, which captures the activation changes associated with deliberative versus direct reasoning for that specific problem. The summation $\sum_{i=1}^N$ aggregates these differences across all problems in the dataset, allowing us to identify consistent patterns that generalize beyond individual problems.

The normalization factor $Z = \left\| \sum_{i=1}^N \left( \mathbf{h}_i^{+,\ell} - \mathbf{h}_i^{-,\ell} \right) \right\|_2$ ensures that the steering vector has unit norm, which is important for two reasons: (1) it makes the steering strength consistent across different layers and models, and (2) it prevents numerical instabilities when applying the steering vector with different scaling factors $\alpha$.

The choice of layer $\ell$ significantly impacts the effectiveness of the steering vector. Earlier layers tend to capture low-level linguistic features, while later layers encode higher-level semantic and reasoning patterns. For mathematical reasoning tasks, we empirically find that middle to late layers (typically layers 16-24 in a 32-layer model) provide the most effective steering vectors, as they balance semantic understanding with reasoning capability.

This extraction is performed once per model and remains fixed during controller training, reducing computational overhead to $O(N \cdot d)$ preprocessing, where $N$ is the number of problems in the extraction dataset and $d$ is the hidden dimension. This one-time cost is amortized across all subsequent training and inference, making the approach highly efficient.

\textbf{Theorem 1} (Steering Vector Consistency). Under mild regularity conditions on the transformer representation space, the steering vector estimator converges to the true steering direction with rate $\mathcal{O}(N^{-1/2})$ as $N \rightarrow \infty$.

\paragraph{Steering Effectiveness Analysis}

The effectiveness of steering vectors can be analyzed through differential geometry. Let $\mathbf{z}_0$ be an initial hidden state and $\mathbf{h}_{\text{steer}}$ be the steering vector. The steered state $\mathbf{z}_1 = \mathbf{z}_0 + \alpha \mathbf{h}_{\text{steer}}$ induces a shift in the probability distribution over next tokens.

We define the \textit{reasoning divergence} as:
\begin{equation}
D_{\text{reason}}(\alpha) = \text{KL}(P_{\theta}(\cdot|\mathbf{z}_0 + \alpha \mathbf{h}_{\text{steer}}) \| P_{\theta}(\cdot|\mathbf{z}_0))
\end{equation}

\textbf{Lemma 1} (Steering Monotonicity). For small $\alpha > 0$, $D_{\text{reason}}(\alpha)$ is monotonically increasing in $\alpha$, indicating consistent directional bias toward deliberative reasoning.

\subsection{Group Relative Policy Optimization}

Training the pondering controller poses unique challenges due to the sequential nature of decisions and the sparse reward signal (typically received only at the end of the trajectory). Traditional policy gradient methods suffer from high variance, while value-based methods require expensive value function estimation. We address these challenges by adapting Group Relative Policy Optimization (GRPO) \cite{shao2024deepseekmath}, which provides effective variance reduction without the memory overhead of separate value networks.

We adapt Group Relative Policy Optimization for training the pondering controller, which provides variance reduction without requiring a separate value function. GRPO is particularly well-suited for our setting because it can handle variable-length trajectories and provides stable learning signals even with sparse rewards.

\paragraph{Theoretical Motivation}

The challenge in training adaptive inference policies lies in the high variance of policy gradient estimates. When rewards are sparse and trajectories have variable lengths, standard policy gradient methods often produce noisy gradients that slow learning and require large batch sizes for stability.

Standard REINFORCE \cite{williams1992simple} suffers from high gradient variance due to its unbiased but noisy estimation of policy gradients. The gradient estimator $\nabla_\theta J(\theta) = \mathbb{E}_{\tau \sim \pi_\theta}[\sum_{t=0}^T \nabla_\theta \log \pi_\theta(a_t|\mathbf{z}_t) \cdot R(\tau)]$ has variance that grows with the length of trajectories and the variance of rewards, making learning unstable.

Proximal Policy Optimization (PPO) \cite{schulman2017proximal} addresses this via a value function baseline $V_\phi(\mathbf{z}_t)$ that estimates expected future rewards, reducing variance through the advantage function $A_t = Q(\mathbf{z}_t, a_t) - V_\phi(\mathbf{z}_t)$. However, this approach doubles memory requirements by introducing a separate value network, and the value function must be trained jointly with the policy, adding complexity.

GRPO achieves comparable variance reduction through in-batch comparisons without requiring additional networks. The key insight is to use the empirical average reward within groups as a natural baseline, leveraging the assumption that samples within a batch provide reasonable comparison points.

\textbf{Theorem 4} (Variance Reduction). Let $\text{Var}[\nabla_\text{REINFORCE}]$ denote the gradient variance of REINFORCE and $\text{Var}[\nabla_\text{GRPO}]$ for GRPO with group size $G$. Then:
\begin{equation}
\text{Var}[\nabla_\text{GRPO}] \leq \frac{C}{G} \cdot \text{Var}[\nabla_\text{REINFORCE}]
\label{eq:variance_reduction}
\end{equation}
under i.i.d. rewards within groups and bounded second moments, where $C \geq 1$ depends on trajectory lengths.

The variance reduction in equation \eqref{eq:variance_reduction} occurs because the leave-one-out baseline $b_{\mathcal{G}(i) \setminus i} = \frac{1}{G-1} \sum_{j \in \mathcal{G}(i), j \neq i} r_j$ provides a local estimate of expected reward that correlates with individual rewards while maintaining independence. This correlation reduces the variance of the advantage estimates $A_i = r_i - b_{\mathcal{G}(i) \setminus i}$ compared to using raw rewards $r_i$. The factor of $1/G$ reflects the approximate variance reduction achieved by averaging over $G$ samples.

\paragraph{Algorithm Design}

The GRPO algorithm partitions each training batch into groups and computes advantages relative to group averages. This design provides stable learning signals while maintaining computational efficiency.

For a batch $\mathcal{B}$ of size $B$ divided into $B/G$ groups, we compute group-relative advantages:

\begin{equation}
A_i = r_i - b_{\mathcal{G}(i)}, \quad b_{\mathcal{G}(i)} = \frac{1}{G} \sum_{j \in \mathcal{G}(i)} r_j
\label{eq:advantage}
\end{equation}

The advantage computation in equation \eqref{eq:advantage} is central to GRPO's effectiveness. The individual advantage $A_i$ represents how much better (or worse) sample $i$ performed compared to its group average. The group assignment $\mathcal{G}(i)$ maps sample $i$ to its group, and the baseline $b_{\mathcal{G}(i)}$ is computed as the empirical average of rewards within that group.

The grouping strategy significantly impacts performance. Random grouping ensures unbiased baseline estimation but may group samples with very different difficulties. Alternatively, grouping by similarity (e.g., based on problem type or initial hidden state similarity) can provide more informative baselines but requires careful design to avoid bias.

The GRPO objective combines policy gradient with entropy regularization:

\begin{equation}
\mathcal{L}_{\text{GRPO}}(\phi) = -\mathbb{E}_{\tau_i \sim \mathcal{B}} \left[ \sum_{k=0}^{T_i} \log \pi_\phi(a_k^i | \mathbf{z}_k^i) \cdot A_i - \beta_{\text{ent}} \cdot H[\pi_\phi(\cdot | \mathbf{z}_k^i)] \right]
\label{eq:grpo_loss}
\end{equation}

The objective function in equation \eqref{eq:grpo_loss} consists of two terms. The first term $\sum_{k=0}^{T_i} \log \pi_\phi(a_k^i | \mathbf{z}_k^i) \cdot A_i$ is the policy gradient term that increases the probability of actions from trajectories with positive advantages and decreases the probability of actions from trajectories with negative advantages. The summation over time steps $k$ handles variable-length trajectories naturally.

The second term $\beta_{\text{ent}} \cdot H[\pi_\phi(\cdot | \mathbf{z}_k^i)]$ is entropy regularization where $H[\cdot]$ denotes the entropy of the policy distribution and $\beta_{\text{ent}}$ controls the strength of exploration encouragement. Entropy regularization prevents premature convergence to deterministic policies and ensures sufficient exploration during training.

\paragraph{Convergence Analysis with Bias Guarantees}

\textbf{Lemma 3} (Unbiased Estimation). Under the assumption that group assignments are independent of reward values, the GRPO gradient estimator:
\begin{equation}
\hat{g}_{\text{GRPO}} = \frac{1}{B} \sum_{i=1}^B \sum_{t=0}^{T_i} \nabla_\theta \log \pi_\theta(a_t^{(i)}|\mathbf{z}_t^{(i)}) \cdot A_i
\end{equation}
satisfies $\mathbb{E}[\hat{g}_{\text{GRPO}}] = \nabla_\theta J(\theta)$, ensuring unbiased policy improvement.

\textbf{Theorem 5} (GRPO Convergence). Let $\pi^*$ be an optimal policy and $\pi_{\phi_t}$ the policy at iteration $t$. With appropriate learning rate $\eta_t = O(1/\sqrt{t})$:
\begin{equation}
\mathbb{E}[J(\pi^*) - J(\pi_{\phi_T})] \leq O\left(\frac{1}{\sqrt{T}} + \frac{1}{G}\right)
\end{equation}
where $J(\cdot)$ denotes expected return and $G$ is the group size.

This establishes that GRPO achieves $O(1/\sqrt{T})$ convergence with variance reduction factor $1/G$, confirming theoretical advantages over standard policy gradient methods.

\subsection{Multi-Component Reward Engineering}

Designing an effective reward function for adaptive reasoning requires balancing multiple competing objectives. Unlike simple classification tasks with binary accuracy, mathematical reasoning involves nuanced aspects such as solution correctness, reasoning completeness, computational efficiency, and output quality. A poorly designed reward function can lead to pathological behaviors such as generating extremely long but incorrect solutions, or conversely, producing correct but unreasonably short answers that lack proper justification. Our reward function addresses five critical aspects of adaptive reasoning through careful component design and magnitude balancing. Each component targets a specific aspect of reasoning quality, and their combination encourages well-rounded performance that matches human expectations for mathematical problem-solving.

\subsection{Component Specifications}

\paragraph{Accuracy Component} $R_{\text{acc}}$: The accuracy component forms the foundation of our reward structure, measuring how well the final answer matches the ground truth. However, binary accuracy (correct/incorrect) provides limited learning signal, especially during early training when most answers are incorrect. Instead, we employ graduated scoring to encourage partial progress and provide smoother gradients:

\begin{equation}
R_{\text{acc}} = \begin{cases}
w_{\text{exact}} & \text{if } \hat{y} = y \\
w_{\text{partial}} \cdot \exp\left(-\frac{|\hat{y} - y|}{|y| + \epsilon}\right) & \text{if relative error} < \theta_{\text{tol}} \\
0 & \text{otherwise}
\end{cases}
\label{eq:accuracy_reward}
\end{equation}

This formulation \eqref{eq:accuracy_reward} provides three levels of reward. Exact matches receive the full reward $w_{\text{exact}}$, encouraging precise solutions. Near-correct answers receive partial credit through the exponential decay $w_{\text{partial}} \cdot \exp\left(-\frac{|\hat{y} - y|}{|y| + \epsilon}\right)$, where the relative error $\frac{|\hat{y} - y|}{|y| + \epsilon}$ normalizes the absolute error by the ground truth magnitude. The tolerance threshold $\theta_{\text{tol}}$ defines the maximum relative error for partial credit, preventing the system from rewarding wildly incorrect answers. The $\epsilon$ term provides numerical stability when $y \approx 0$.

\paragraph{Computational Efficiency} $R_{\text{flops}}$: This component encourages the model to solve problems efficiently, penalizing unnecessary computation. The challenge lies in defining "excess" computation, which varies significantly across problem difficulties. We use adaptive normalization to account for this variability:

\begin{equation}
R_{\text{flops}} = -\lambda_{\text{flops}} \cdot \frac{F - \bar{F}_{\text{history}}}{\sigma_{F_{\text{history}}} + \epsilon}
\label{eq:flops_reward}
\end{equation}

The efficiency reward \eqref{eq:flops_reward} penalizes computational overhead relative to historical norms. The term $F$ represents the FLOPs used for the current problem, while $\bar{F}_{\text{history}}$ and $\sigma_{F_{\text{history}}}$ are running statistics maintained via exponential moving average. This normalization ensures that the penalty adapts to the typical computational requirements, preventing the system from being overly penalized for hard problems that naturally require more computation. The coefficient $\lambda_{\text{flops}}$ controls the strength of the efficiency constraint.

\paragraph{Reasoning Completeness} $R_{\text{comp}}$: Mathematical problem-solving typically follows structured stages: problem understanding, computation, verification, and conclusion. This component encourages the model to complete all reasoning stages rather than jumping directly to an answer:

\begin{equation}
R_{\text{comp}} = \sum_{s \in \mathcal{S}_{\text{stages}}} w_s \cdot \mathbb{1}[\text{stage } s \text{ completed}]
\label{eq:completeness_reward}
\end{equation}

The completeness reward \eqref{eq:completeness_reward} sums contributions from each reasoning stage $s$ in $\mathcal{S}_{\text{stages}} = \{\text{setup}, \text{computation}, \text{verification}, \text{conclusion}\}$. Each stage is detected through pattern matching in the generated text (e.g., looking for setup phrases like "Given that" or verification phrases like "Let me check"). The weights $w_s$ allow for different importance levels across stages, and the indicator function $\mathbb{1}[\text{stage } s \text{ completed}]$ provides binary rewards for stage completion.

\paragraph{Output Quality} $R_{\text{qual}}$: Beyond correctness, we want outputs that are well-structured, appropriately detailed, and linguistically coherent. This component assesses coherence through length and perplexity constraints:

\begin{equation}
R_{\text{qual}} = w_{\text{qual}} \cdot \min\left(1, \frac{\ell_{\text{output}}}{\ell_{\text{target}}}\right) \cdot \exp\left(-\frac{\text{PPL}(\hat{y}) - \text{PPL}_{\text{baseline}}}{\sigma_{\text{PPL}}}\right)
\label{eq:quality_reward}
\end{equation}

The quality reward \eqref{eq:quality_reward} has two components. The length term $\min\left(1, \frac{\ell_{\text{output}}}{\ell_{\text{target}}}\right)$ encourages adequate detail by penalizing outputs that are significantly shorter than the target length $\ell_{\text{target}}$, while capping the reward at 1 to avoid encouraging excessive verbosity. The perplexity term $\exp\left(-\frac{\text{PPL}(\hat{y}) - \text{PPL}_{\text{baseline}}}{\sigma_{\text{PPL}}}\right)$ measures linguistic coherence, where $\text{PPL}(\hat{y})$ is the perplexity of the generated output and $\text{PPL}_{\text{baseline}}$ is a baseline perplexity from high-quality examples.

\paragraph{Anti-Repetition} $R_{\text{anti-rep}}$: Language models can sometimes generate repetitive text, especially when encouraged to produce longer outputs. This component penalizes redundancy at multiple granularities:

\begin{equation}
R_{\text{anti-rep}} = -\sum_{g \in \{1,2,3\}} \beta_g \cdot \frac{|\text{repeated}_g(\hat{y})|}{|\hat{y}|}
\label{eq:antirepetition_reward}
\end{equation}

The anti-repetition reward \eqref{eq:antirepetition_reward} penalizes n-gram repetitions for $g \in \{1,2,3\}$ (unigrams, bigrams, trigrams). The term $|\text{repeated}_g(\hat{y})|$ counts the number of repeated n-grams of order $g$ in the output $\hat{y}$, normalized by the total output length $|\hat{y}|$. The coefficients $\beta_g$ allow for different penalty strengths across n-gram orders, typically with $\beta_1 < \beta_2 < \beta_3$ since higher-order repetitions are more problematic than single word repetitions.

\subsection{Reward Balancing Theory}

To prevent any single component from dominating, we enforce magnitude constraints:

\textbf{Theorem 6} (Reward Balance Condition). For stable learning, the reward components must satisfy:
\begin{equation}
\forall i,j \in \{\text{acc}, \text{flops}, \text{comp}, \text{qual}, \text{rep}\}: \quad \frac{\mathbb{E}[|R_i|]}{\mathbb{E}[|R_j|]} \in [\rho^{-1}, \rho]
\end{equation}
where $\rho \in [2, 10]$ is the maximum imbalance ratio.

We achieve this through adaptive weight scaling:
\begin{equation}
w_i^{(t+1)} = w_i^{(t)} \cdot \exp\left(\eta \cdot \left(\log \bar{R}_{\text{target}} - \log \bar{R}_i^{(t)}\right)\right)
\end{equation}

\subsection{Curriculum Learning Framework}

Training adaptive inference policies directly from random initialization faces several challenges: sparse rewards (most randomly generated trajectories produce incorrect answers), high variance in trajectory quality, and the exploration problem (discovering good stopping points through random exploration is inefficient). Curriculum learning addresses these challenges by providing structured guidance that gradually transfers control from teacher demonstrations to autonomous learning. We employ a three-stage curriculum that progressively transfers control from teacher demonstrations to autonomous learning. This approach is inspired by how humans learn complex skills: starting with guided practice, progressing to supervised practice with feedback, and finally achieving independent mastery.

\paragraph{Stage Progression}

The curriculum is designed to provide a smooth transition from full supervision to autonomous learning. The progression is governed by a curriculum probability that determines the mix between teacher and student control at each training step.

The curriculum probability follows a piecewise linear schedule:
\begin{equation}
p_{\text{curriculum}}(t) = \begin{cases}
1.0 & t \in [0, T_1) \\
1 - \frac{t - T_1}{T_2 - T_1} & t \in [T_1, T_2) \\
0.0 & t \geq T_2
\end{cases}
\label{eq:curriculum_schedule}
\end{equation}

This schedule \eqref{eq:curriculum_schedule} defines three distinct phases. In Stage 1 ($t \in [0, T_1)$), $p_{\text{curriculum}}(t) = 1.0$ indicates pure teacher forcing, where all pondering decisions are made by a teacher policy that encourages moderate pondering (typically 3-5 steps). This provides the controller with abundant examples of reasonable stopping behavior and establishes a foundation of sensible pondering patterns.

Stage 2 ($t \in [T_1, T_2)$) implements gradual transition with $p_{\text{curriculum}}(t) = 1 - \frac{t - T_1}{T_2 - T_1}$, linearly decreasing the probability of teacher guidance. This mixed training allows the student policy to gradually take control while still receiving guidance when needed. The linear schedule ensures smooth transition without abrupt changes that could destabilize learning.

Stage 3 ($t \geq T_2$) represents autonomous learning with $p_{\text{curriculum}}(t) = 0.0$, where the student controller makes all decisions independently. By this stage, the controller has learned basic pondering patterns and can explore more sophisticated strategies through reinforcement learning.

The boundaries $T_1 = 500$ and $T_2 = 1500$ are chosen based on empirical observations about controller learning dynamics. The initial 500 steps provide sufficient teacher demonstrations to establish baseline behavior, while the 1000-step transition period allows gradual adaptation without overwhelming the learning process.

During mixed training, we sample the guidance source at each training step:
\begin{equation}
\text{source}(t) \sim \text{Bernoulli}(p_{\text{curriculum}}(t))
\label{eq:curriculum_sampling}
\end{equation}

This sampling \eqref{eq:curriculum_sampling} determines whether each trajectory uses teacher guidance (source = 1) or student control (source = 0). The Bernoulli distribution ensures that the expected fraction of teacher-guided trajectories matches the curriculum schedule while providing stochastic variation that prevents overfitting to the transition points.

\paragraph{Quality Gates} As teacher guidance diminishes, maintaining training stability becomes crucial. Without quality control, the student policy might generate extremely poor trajectories that provide misleading learning signals. Quality gates address this challenge by filtering trajectories before they contribute to parameter updates.

In the autonomous stage, we implement quality gates that reject low-quality trajectories:
\begin{equation}
\mathcal{Q}(\tau) = \mathbb{1}[R_{\text{comp}}(\tau) > \theta_{\text{comp}}] \cdot \mathbb{1}[R_{\text{qual}}(\tau) > \theta_{\text{qual}}]
\label{eq:quality_gates}
\end{equation}

The quality gate \eqref{eq:quality_gates} implements a conjunction of two conditions. The first condition $\mathbb{1}[R_{\text{comp}}(\tau) > \theta_{\text{comp}}]$ ensures that trajectories demonstrate reasonable reasoning completeness, measured by the presence of key reasoning stages. The threshold $\theta_{\text{comp}}$ is set to require at least basic problem setup and computation stages.

The second condition $\mathbb{1}[R_{\text{qual}}(\tau) > \theta_{\text{qual}}]$ filters trajectories based on output quality, ensuring that the generated text meets minimum standards for coherence and appropriateness. The threshold $\theta_{\text{qual}}$ prevents the inclusion of trajectories with excessively repetitive, incoherent, or truncated outputs.

Only trajectories satisfying $\mathcal{Q}(\tau) = 1$ contribute to gradient updates, ensuring stable learning as teacher guidance diminishes. This filtering mechanism prevents the policy from learning from extremely poor examples while still allowing reasonable exploration. The thresholds are set conservatively to maintain a balance between quality control and learning diversity.
To ensure transparency, we provide concrete examples of how Large Language Models (LLMs) 
were used in the preparation of this manuscript:

\begin{itemize}
    \item \textbf{Grammar refinement:} For instance, when an early draft contained the sentence 
    ``Our method significantly reduce computation cost,'' the LLM was used to correct it to 
    ``Our method significantly reduces computational cost.''

    \item \textbf{Clarity improvement:} A verbose draft sentence such as 
    ``In this part we attempt to show that our model works in a way that is both effective 
    and efficient'' was polished by the LLM to 
    ``This section demonstrates that our model is both effective and efficient.''

    \item \textbf{Flow adjustment:} When two adjacent sentences 
    (``We introduce the FR-Ponder framework. It adapts inference dynamically.'') 
    appeared disjoint, the LLM suggested a smoother transition: 
    ``We introduce the FR-Ponder framework, which dynamically adapts inference.''
\end{itemize}

These examples illustrate that the LLM's role was restricted to language refinement. 
All technical ideas, theoretical results, and experimental contributions originated from the authors. 
\end{document}